\def\year{2022}\relax
\documentclass[letterpaper]{article} % DO NOT CHANGE THIS
\usepackage{aaai22}  % DO NOT CHANGE THIS
\usepackage{times}  % DO NOT CHANGE THIS
\usepackage{helvet}  % DO NOT CHANGE THIS
\usepackage{courier}  % DO NOT CHANGE THIS
\usepackage[hyphens]{url}  % DO NOT CHANGE THIS
\usepackage{graphicx} % DO NOT CHANGE THIS
\usepackage{natbib}  % DO NOT CHANGE THIS AND DO NOT ADD ANY OPTIONS TO IT
\usepackage{caption} % DO NOT CHANGE THIS AND DO NOT ADD ANY OPTIONS TO IT
\DeclareCaptionStyle{ruled}{labelfont=normalfont,labelsep=colon,strut=off} % DO NOT CHANGE THIS
\usepackage{algorithm}
\usepackage{algorithmic}
\usepackage{amssymb}
\usepackage{booktabs}
\usepackage{amsmath}
\usepackage{multirow}
\usepackage{bbding}
\usepackage{mathrsfs}
\usepackage{subfigure}
\usepackage{newfloat}
\usepackage{listings}
\floatstyle{ruled}
\newfloat{listing}{tb}{lst}{}
\floatname{listing}{Listing}

\title{TODSum: Task-Oriented Dialogue Summarization with State Tracking}
\author{Lulu Zhao$^{1}$, Fujia Zheng$^{1}$, Keqing He$^{1}$, Weihao Zeng$^{1}$, Yuejie Lei$^{1}$, Huixing Jiang$^{2}$, Wei Wu$^{2}$, Weiran Xu$^{1}$\thanks{\ \ Weiran Xu is the corresponding author.}, Jun Guo$^{1}$, Fanyu Meng$^{3}$\\
  $^1$Beijing University of Posts and Telecommunications, Beijing, China\\
  $^2$Meituan, Beijing, China\\
  $^3$China Mobile Communication Corporation, Beijing, China\\
  \texttt{\{zhaoll,fujia\_zheng,kqin,ZengWH,2019110830,xuweiran,guojun\}@bupt.edu.cn}
  \texttt{\{jianghuixing\}@meituan.com}
  \texttt{\{wuwei19850318\}@gmail.com}
  \texttt{\{mfybupt2008\}@163.com}
  }

\usepackage{bibentry}
\begin{document}

\maketitle

\begin{abstract}
Previous dialogue summarization datasets mainly focus on open-domain chitchat dialogues, while summarization datasets for the broadly used task-oriented dialogue haven't been explored yet. Automatically summarizing such task-oriented dialogues can help a business collect and review needs to improve the service. Besides, previous datasets pay more attention to generate good summaries with higher ROUGE scores, but they hardly understand the structured information of dialogues and ignore the factuality of summaries. In this paper, we introduce a large-scale public Task-Oriented Dialogue Summarization dataset, TODSum, which aims to summarize the key points of the agent completing certain tasks with the user. Compared to existing work, TODSum suffers from severe scattered information issues and requires strict factual consistency, which makes it hard to directly apply recent dialogue summarization models. Therefore, we introduce additional dialogue state knowledge for TODSum to enhance the faithfulness of generated summaries. We hope a better understanding of conversational content helps summarization models generate concise and coherent summaries. Meanwhile, we establish a comprehensive benchmark for TODSum and propose a state-aware structured dialogue summarization model to integrate dialogue state information and dialogue history. Exhaustive experiments and qualitative analysis prove the effectiveness of dialogue structure guidance. Finally, we discuss the current issues of TODSum and potential development directions for future work.
\end{abstract}

\section{Introduction}

% 1. 简要介绍摘要任务，引出对话摘要定义。
% 2. 按照数据集类别，介绍之前的摘要数据集工作，引出我们的任务型对话摘要，突出与之前数据集的区别。
% 3. 讨论任务型对话摘要面临的挑战（补充个示意图），突出我们研究问题的重要性。（注意从case study入手，相互对齐）
% 4. 介绍本文主要内容
    % 1. 定义任务
    % 2. 构造数据集
    % 3. 提出指标
    % 4. 提出新模型
    % 5. 实验提升
    % 6. 分析论证
% 5. Contribution
    % 1. 定义了任务型对话摘要任务，并提出了一个新的结合对话状态的摘要数据集。
    % 2. 提出了一种基于对话结构信息增强的摘要模型，有效提升传统摘要模型的事实一致性，并建立了一个benchmark供后续研究。
    % 3. 分析了任务型对话摘要任务面临的问题与挑战，以及引入对话结构信息带来的提升

% 对话结构的意义

% 从 Faithfulness 和 Controllability入手，对于一个客服/任务型对话系统来说，要优先保证生成摘要的事实一致性，而引入对话状态可以增加现有摘要模型的可控性，保证生成摘要包含对话中关键的意图和槽位信息。
% 2. 从对话建模的角度去理解摘要系统，定义新指标；对话理解和摘要相互促进（联合模型）
\begin{figure}[h!]
\centering{
\includegraphics[scale=0.7]{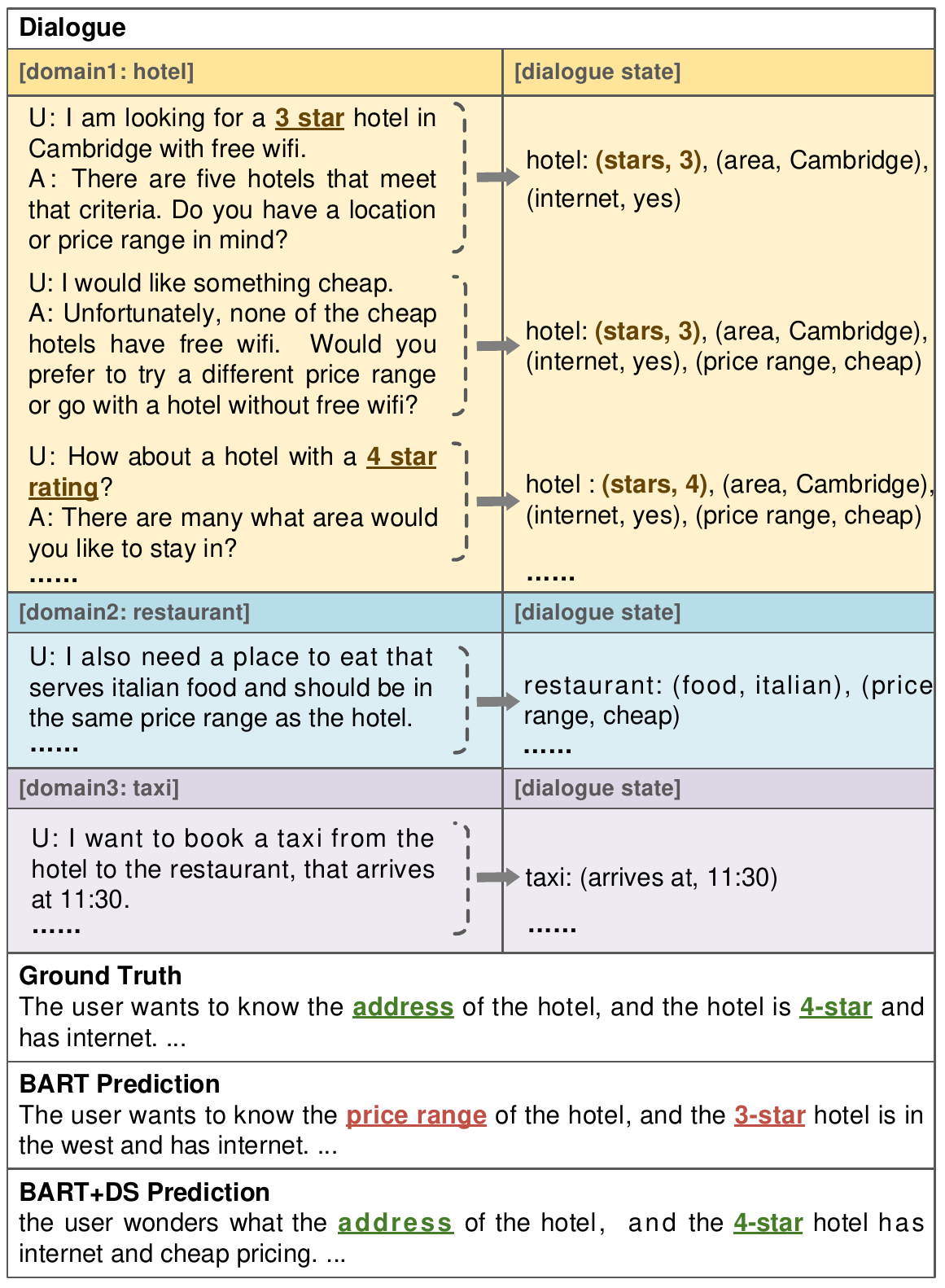}}
\vspace{-0.3cm}
\caption{An example of TODSum with dialogue states. There exists three challenges: \textbf{Factual Inconsistency} (factual errors in generated summaries like \emph{3-star}), \textbf{Repetition and Negotiation} (see \emph{hotel}), and \textbf{Multiple Domains}.}
\label{fig:intro_case}
\vspace{-0.722cm}
\end{figure}

Text summarization is the task of automatically generating a concise, salient, coherent, and fluent summary of given long text input \cite{Radev2002IntroductionTT}. Recent work in text summarization has made significant progress in news articles \cite{Nallapati2016AbstractiveTS,Kedzie2018ContentSI}, scientific papers \cite{KoncelKedziorski2019TextGF}, and patents \cite{Sharma2019BIGPATENTAL}. Furthermore, the large pre-trained models, such as BERT \cite{Devlin2019BERTPO}, BART \cite{Lewis2020BARTDS}, and Pegasus \cite{Zhang2020PEGASUSPW} boost the performance.
% , especially in zero and few-shot settings \cite{Fabbri2021ImprovingZA}
However, dialogue summarization receives significantly less attention mainly attributed to the complex context, multiple speakers, and informal language style \cite{Chen2020MultiViewSM}. Therefore, more valuable conversational datasets for summarization are required to facilitate the related research area.

% According to different conversation scenarios, previous dialogue summarization datasets are generally classified into open-domain chitchat \cite{Gliwa2019SAMSumCA, Malykh2020SumTitlesAS, Chen2021DialogSumAR, Fabbri2021ConvoSummCS}, meeting \cite{Carletta2005TheAM, Zhong2021QMSumAN}, multi-turn question answering \cite{Chowdhury2019CQASUMMBR} and customer-service dialogues \cite{Liu2019AutomaticDS, Yuan2019AbstractiveDS, Zou2021UnsupervisedSF, Zou2021TopicOrientedSD}.  Open-domain chitchat, like SAMSum \cite{Gliwa2019SAMSumCA} are mostly used which contains human-written online social chats. These chats typically are diverse rather short. Then several spoken daily dialogue summarization datasets are proposed such as DIALSUMM \cite{Chen2021DialogSumAR} and ConvoSumm \cite{Fabbri2021ConvoSummCS}, which contains discussion of a specific topic like the Super Bowl, travel, etc. Meeting summarization such as AMI \cite{Carletta2005TheAM} and QMSum \cite{Zhong2021QMSumAN} focus on covering all the key points of a long meeting involving multiple people and topics. Multi-turn QA summarization like CQASumm \cite{Chowdhury2019CQASUMMBR} instead summarize user question and system answer, such as Quora and Stackoverflow. 
According to different conversation scenarios, previous dialogue summarization datasets are generally classified into the open-domain chitchat, meeting, multi-turn question answering, and customer-service dialogues. Open-domain chitchat, like SAMSum \cite{Gliwa2019SAMSumCA} and DialSumm \cite{Chen2021DialogSumAR}, are mostly used which contains human-written online social chats. 
% These chats typically are diverse rather short. 
Meeting such as AMI \cite{Carletta2005TheAM}, MediaSum \cite{zhu-etal-2021-mediasum}, and QMSum \cite{Zhong2021QMSumAN} focus on the meeting scenario involving multiple people and topics. Multi-turn QA like CQASumm \cite{Chowdhury2019CQASUMMBR} and ConvoSumm \cite{Fabbri2021ConvoSummCS} instead summarize user questions and system answers like Quora. 
% Customer-service dialogues \cite{Liu2019AutomaticDS, Yuan2019AbstractiveDS, Zou2021UnsupervisedSF, Zou2021TopicOrientedSD} are recently proposed to address user issues about specific topics, such as E-commerce after-sales service or counseling. These conversations usually contain concrete motivations and unified responses. However, these above datasets always omit the intrinsical dialogue state or structure of multi-turn conversation and simply regard dialogue history as long text \cite{Liu2019TextSW}. Several work \cite{Chen2020MultiViewSM, Chen2021StructureAwareAC} try to construct dialogue stage or discourse graph or action graph from raw dialogue history. We argue this information can't reveal major characteristics of multi-turn dialogues, especially task-oriented dialogues which contain domain-specific structured ontology like intents and slots. Therefore, in this paper, we propose a novel \textbf{T}ask-\textbf{O}riented \textbf{D}ialogue \textbf{Sum}marization dataset, TODSum, with corresponding dialogue state knowledge. We hope these explicit dialogue structures can enhance the faithfulness and controllability of generated summaries.
Customer-service dialogues \cite{Liu2019AutomaticDS, Yuan2019AbstractiveDS, Zou2021UnsupervisedSF, Zou2021TopicOrientedSD} are recently proposed to address user issues about specific topics, such as the E-commerce after-sales service. But due to privacy limitations, existing customer-service dialogues are not publicly available for research. 

Overall, most existing dialogue summarization datasets focus on open chitchats but overlook another kind of important dialogue type, the Task-Oriented Dialogue (TOD) \cite{Zhang2020RecentAA} which is widely used in practical scenarios.
To the best of our knowledge, there is no existing large-scale public summarization dataset for TOD. More importantly, although previous datasets make some progress, a critical question is that all of the datasets ignore the intrinsical dialogue structured information of multi-turn conversations and only focus on summarization generation, like ROUGE scores and Bleu, rather than dialogue understanding, i.e., intents and  slots. Several works \cite{Chen2020MultiViewSM, Chen2021StructureAwareAC} try to extract dialogue topics, discourse, or action graphs from raw dialogue history using handcrafted rules or unsupervised methods. However, this information can't reveal essential characteristics of multi-turn dialogues, especially the task-oriented dialogues which contain domain-specific structured ontologies like intents and slots \cite{he-etal-2020-learning-tag,yan-etal-2020-adversarial}. Therefore, in this paper, we propose a large-scale \textbf{T}ask-\textbf{O}riented \textbf{D}ialogue \textbf{Sum}marization dataset, TODSum with dialogue state knowledge to summarize the key points of the agent completing certain tasks with the user, such as the user's target(intent), preference (cheap or expensive restaurant), and user questions. Automatically summarizing such task-oriented dialogues can help a business collect and review needs or complaints from customers to improve the service. We hope to provide a unified and high-quality structured dialogue summarization benchmark to facilitate dialogue understanding while generating summaries. 

Fig \ref{fig:intro_case} displays an example of TODSum where a user wants to make a reservation. Compared to existing datasets, TODSum faces the following key challenges: (1) \textbf{Factual Inconsistency}: Multiple slot values often confuse summarization models to generate summaries with factual errors, like \emph{3-star} vs \emph{4-star}. (2) \textbf{Repetition and Negotiation}: The user usually negotiates some slot values with the agent many times since there is no available reservation. For example, the user changes \emph{3-star} to \emph{4-star} since there is no available hotel meeting all the constraints. (3) \textbf{Multiple Domains}: TODSum contains multiple domains among a single dialogue like \emph{hotel}, \emph{restaurant} and, \emph{taxi}.

To tackle these challenges, we introduce additional dialogue state information for TODSum to enhance the faithfulness and controllability of generated summaries. We hope the better understanding of conversational content helps summarization models focus on key information and generate concise and coherent summaries. Considering that TODSum pays more attention to the factual consistency of generated summaries, we propose new state-aware factual consistency evaluation metrics based on the annotated dialogue state information for exhibiting the faithfulness of models. Then, we establish a fair benchmark and extensive strong dialogue summarization baselines for TODSum. Besides, we also propose an efficient state-aware structured dialogue summarization model to integrate the dialogue history and state information. Exhaustive experiments and analysis prove the effectiveness of dialogue structure guidance for task-oriented dialogue summarization. Finally, we perform a qualitative analysis to shed light on the current issues and future directions for TODSum. Our contributions are three-fold: (1) We introduce a task-oriented dialogue summarization dataset, TODSum along with corresponding dialogue state knowledge. (2) We establish a comprehensive benchmark and propose a general dialogue structure-aware summarization model to combine the original dialogue text and structured dialogue state. Besides, we propose new state-aware factual consistency metrics. (3) We conduct exhaustive qualitative analysis to prove the effectiveness of dialogue structure guidance and discuss current issues of TODSum and potential development directions for future work.

\section{Related Work}
\noindent\textbf{Abstractive Dialogue Summarization Datasets}
Recently, many dialogue summarization datasets covering various scenarios have been constructed. Some research directly adopts meeting and interview transcripts as a special kind of dialogic text, such as AMI \cite{Carletta2005TheAM}, ICSI \cite{1198793}, and MediaSum \cite{zhu-etal-2021-mediasum}. Since the meeting is essentially different from the dialogue in the language pattern, researchers further propose some practical conversational scenario datasets. They are either about the open-domain chitchats \cite{Gliwa2019SAMSumCA,Chen2021DialogSumAR} or the multi/single turn QA scenarios \cite{Chowdhury2019CQASUMMBR,Fabbri2021ConvoSummCS}. Additionally, the studies on the customer-service dialogue summarization datasets, which is about the after-sales service or counseling, have attracted more attention and they belong to a task-oriented dialogue system. \citet{Liu2019AutomaticDS} proposed a large-scale dataset from the logs in the DiDi customer service center and \citet{Zou2021UnsupervisedSF,Zou2021TopicOrientedSD} collected a call-log dataset from the  E-commerce platform. Both of them are faced with privacy issues and have higher labeling costs due to the human-written summaries. \citet{Yuan2019AbstractiveDS} used the MultiWOZ dataset and use the self-configured instructions as the golden summaries. Because of the too-long length and the low quality of instructions, they are not suitable for choosing as the gold summaries. It is worth noting that although previous dialogue datasets have achieved some success, they ignore the most important characteristics of task-oriented dialogue, i.e., dialogue state structures. Therefore, to fill in this gap, we propose a novel TODSum dataset, which emphasizes the unique dialogue state structures of task-oriented dialogues.

\noindent\textbf{Factual Consistency in Summarization}
For the fact fabrication issue, some researchers are committed to designing evaluation metrics towards factual consistency, which is because that ROUGE scores correlate poorly with faithfulness. They are divided into 4 types: triple-based \cite{10.1145/3292500.3330955,zhang-etal-2020-optimizing}, textual entailment-based \cite{falke-etal-2019-ranking}, QA-based \cite{wang-etal-2020-asking,durmus-etal-2020-feqa}, and pre-trained classifiers-based \cite{kryscinski-etal-2020-evaluating}. Another line of the related work focuses on enforcing factual relations in summarization models. \citet{cao2017faithful,zhu-etal-2021-enhancing} proposed to encode facts in the sequential way and graph-based way, respectively. \citet{li-etal-2018-ensure} proposed an entailment-reward augmented maximum-likelihood training objective. \citet{dong-etal-2020-multi,cao-etal-2020-factual} designed post-editing correctors to boost factual consistency in generated summaries. Considering that slots and intents are the two most important components in the task-oriented dialogue system, once there are a large number of factual errors about them appearing in the generated summaries, the summary system is completely unusable. In this paper, we try to boost the factual consistency via incorporating the dialogue state information while generating the summaries, and we also design novel state-aware factual evaluation metrics.

\noindent\textbf{Task-Oriented Dialogue System} Compared to open-domain chitchats whose goal is to maximize user engagement\cite{Huang2020ChallengesIB}, a task-oriented dialogue(TOD) system aims to assist the user in completing certain tasks in a specific domain, such as restaurant booking, which is valuable for the real-world business \cite{Zhang2020RecentAA}. TOD systems are built on top of a structured ontology, which defines the domain knowledge of the tasks, like intents and slots. A key component of TOD is the dialogue state tracker (DST) which tracks the dialogue process in each time step by taking the entire dialog context as input \cite{wu2019transferable,Wang2020DialogueST}.

\section{Problem Formulation}
% keqing
Given a dialogue $\mathbf{x} = \{x_{1}, ..., x_{m}\}$ with $m$ utterances, traditonal dialogue summarization datasets aim to generate the target summary $\mathbf{y}= \{y_{1}, ..., y_{k}\}$ conditioned on the source dialogue $\mathbf{x}$ as $\log p\left(\mathbf{y}^{i} \mid \mathbf{x}^{i} ; \theta\right)$,
% $$
% \arg \max _{\theta} \sum_{\left\langle\mathbf{x}^{i}, \mathbf{y}^{i}\right\rangle \in\langle\mathcal{X}, \mathcal{Y}\rangle} \log p\left(\mathbf{y}^{i} \mid \mathbf{x}^{i} ; \theta\right)
% $$
where $\theta$ are model parameters. Furthermore, TODSum feeds the dialogue state $\mathbf{s}$ into the model in addition to the source dialogue $\mathbf{x}$:
$$
\arg \max _{\theta} \sum_{\left\langle\mathbf{x}^{i}, \mathbf{y}^{i}, \mathbf{s}^{i}\right\rangle \in\langle\mathcal{X}, \mathcal{Y}, \mathcal{S}\rangle} \log p\left(\mathbf{y}^{i} \mid \mathbf{x}^{i}, \mathbf{s}^{i} ; \theta\right)
$$
where the dialogue state $\mathbf{s}$ is a set of slots and their corresponding values extracted from the conversation, such as (price, cheap) and (area, center). Note that TODSum contains multiple domains and intents thus we also add the domain and intent fields of a given pair, i.e., (slot, value), to avoid overlapping.

\section{TODSum}

\begin{table*}[]
\centering
\resizebox{0.9\textwidth}{!}{%
\begin{tabular}{l|cccccccc}
\hline
\textbf{Dataset}    & \textbf{Size}    & \textbf{Dialog.len} & \textbf{Turn.num} & \textbf{Summ.len} & \textbf{Lang} & \textbf{Multi-domain} & \textbf{Public} \\ \hline
AMI \cite{Carletta2005TheAM}       & 137    & 6000.7     & 535.6            & 296.6    & EN   & \XSolidBrush            & \Checkmark       \\
SAMSum \cite{Gliwa2019SAMSumCA}     & 16,369  & 83.9       & 9.9            & 20.3     & EN   & -            & \Checkmark      \\
CQASumm \cite{Chowdhury2019CQASUMMBR}   & 100,001 & 781.4     & 12                & 100      & EN   & \XSolidBrush         &  \Checkmark      \\
E-commerce \cite{zou2020topic} & 18,860  & 1285.3    & 26.1           & 54.2    & CH   & \XSolidBrush            & \XSolidBrush      \\
DialSumm \cite{Chen2021DialogSumAR} & 13,640 & 131        & 9.5              & 22.6     & EN   & -            & \Checkmark      \\
QMSum \cite{Zhong2021QMSumAN}     & 232    & 9069.8     & 556.8          & 69.6     & EN   & \Checkmark             & \Checkmark    \\
% MEDIASUM \cite{zhu-etal-2021-mediasum}  & 463,596 & 1553.7     & 30.0     & 6.5      & 14.4     & EN   & \Checkmark            & \Checkmark      \\

ConvoSumm \cite{Fabbri2021ConvoSummCS} & 2,021   & 1,096      & 9.8            & 72.8     & EN   & \XSolidBrush          & \Checkmark       \\
TODSum(ours)     & 9,906   & 186.9     & 14.1              & 45.4     & EN   & \Checkmark          & \Checkmark       \\ \hline
\end{tabular}%
}
\vspace{-0.2cm}
\caption{Comparison between TODSum and existing dialogue summarization datasets. Summ.len denotes the average length of golden summaries. Multi-domain denotes whether the dataset contains multi-domain cases among a single conversation.}
\label{tab:dataset}
\vspace{-0.5cm}
\end{table*}

% In view of task-oriented dialogue datasets are widely used, we take the advantage of MultiWOZ 2.0 dataset \cite{budzianowski2018multiwoz} to build our task-oriented dialogue summarization dataset TODSum. We first perform a small amount of manual annotation, then use a few-shot task-oriented dialogue generation model SC-GPT \cite{peng2020few} to automatically generate the reference summary, and finally execute a series of post-processing.
%这一块步骤是不是可以改一下？

In this section, we introduce our dataset selection, construction and processing, the characteristics of our proposed dataset TODSum, and new state-aware factual consistency evaluation metrics. 
% We select the widely used MultiWOZ \cite{budzianowski2018multiwoz} as our task-oriented dialogue dataset and construct summarization annotations using human-in-the-loop interaction. Then we perform the data post-processing and report the statistics of TODSum compared to existing datasets. Finally, we propose new factual consistency evaluation metrics, state-aware Precision/Recall/F1.

\subsection{Data Collection}
We construct the TODSum based on MultiWOZ \cite{budzianowski2018multiwoz}, which is the largest existing human conversational corpus containing 10,438 samples over 7 domains. MultiWOZ has 30 \emph{(domain, slot)} pairs and over 4,500 possible values. Following \citet{wu2019transferable}, we use five domains \emph{(restaurant, hotel, attraction, taxi, train)} because the other two domains \emph{(hospital, police)} lack enough annotations.

\subsection{Dataset Construction and Processing}

\begin{figure}[t]
\centering{
\includegraphics[scale=0.7]{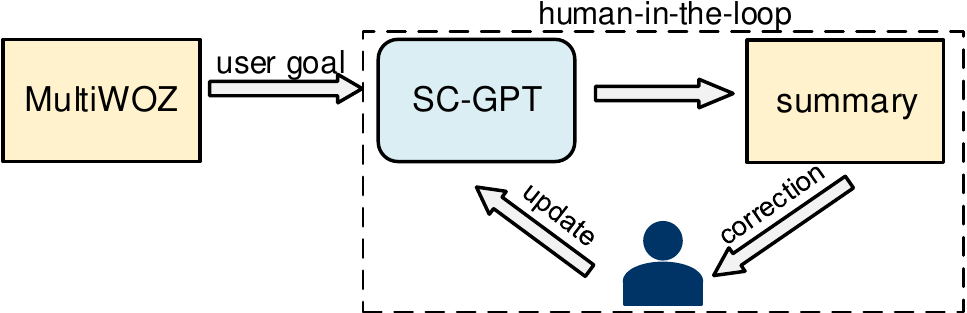}}
\caption{Human-in-the-loop TODSum construction.}
\label{fig:dataset_process}
\vspace{-0.6cm}
\end{figure}

% In this part, we specifically describe the process of dataset construction, including manual annotation, automatic model annotation, and multiple loops of the two.
In this section, we will introduce our annotation method using the human-in-the-loop interaction. The overall procedure is shown in Fig \ref{fig:dataset_process}. Firstly, we use the user goals from MultiWOZ as input and feed them into the state-of-the-art natural language generation model SC-GPT \cite{peng2020few} to generate candidate summaries. Then we perform the human evaluation and correction on a part of generated summaries. The corrected summaries can be used to re-train SC-GPT. We iteratively repeat the above process and finally obtain a well-annotated TODSum.

\textbf{Automatic Annotation} To avoid expensive human annotations, we use the SC-GPT to generate candidate summaries from user goals related to each dialogue. MultiWOZ uses a Wizard-of-Oz framework (WOZ) of human-to-human data collection, where each dialogue is set up with a corresponding user goal. The user goal can be regarded as a task template containing all necessary key information of the dialogue, such as intents and slots. We display an example in the Appendix. Then we use a strong Template2Text model SC-GPT to generate summaries. To control the quality of summaries, we also introduce human correction.
% TODO：这里补充一个例子

\textbf{Human Evaluation and Correction} We randomly sample 100 generated \emph{template, summary} for each domain and modify the generated summaries. Then we finetune the SC-GPT using human-annotated labels. 

\textbf{Human-in-the-loop} We perform the human-in-the-loop interaction to improve the quality of generated summaries without excessive labor-consuming annotations. We interactively perform the automatic annotation and human correction three times. Finally, we obtain the high-quality TODSum with 95\% accuracy \footnote{We employ three annotators to independently evaluate whether the summary is suitable for the raw dialogue. We think the summary is accurate only when all the annotators say yes.}.

% \textbf{Data Post-processing} We collect all the data generated, clean and filter, and divide the dataset according to the division method of \citet{wu2019transferable}. At the same time, we take out the dialogue state of the last round of the dialogue, as the final state of the dialogue, it represents the state of the entire dialogue.

% Note that the user goal we use to annotate the summary is different from the above-mentioned dialogue state. The user goal represents the user's purpose, and we hope that our summary is user-oriented. The dialogue state represents the final state of the dialogue, such as the booking status and information of a restaurant. This information can guide the model to better capture the key information in the dialogue.

\textbf{Data Post-processing} We clean the annotated data and divide training/valid/test following \citet{wu2019transferable}. Note that we also keep the original dialogue state annotations of MultiWOZ. We hope this information can guide the summarization model to better model the dialogue context and generate faithful summaries. Besides, we introduce two dataset variants, Noisy-TODSum and DA-TODSum, where the former adds noise to the golden dialogue state annotations to verify the model robustness, and the latter focuses on the model generalizability on domain adaptation. We leave the details to the related analysis sections.

\subsection{Statistic and Analysis}
% Table \ref{tab:dataset} shows some details of current datasets and our TODSum.

% On the basis of our original TODSum, we also create other variants. 
% One is that we use the classic DST model TRADE \cite{wu2019transferable} to predict the corresponding dialogue state for the test set as part of the pipeline model we mentioned later. 
% The second is that make some disturbances to the dialogue state of the test set and create multiple variants. In the testing phase, the disturbed state information will affect the performance of the model, which also proves the importance of dialogue structure information.
% The third is that we have divide our TODSum into domains to test the domain transfer ability of our subsequent models.

Table \ref{tab:dataset} shows the statistics comparison between TODSum and existing dialogue summarization datasets. TODSum contains 1.9 domains, 3.7 intents, and 12 slots for each dialogue. Multi-domain and complex negotiation increase the difficulty of TODSum. We can find open-domain datasets, i.e., SAMSum and DialSumm, have fewer turns and shorter dialogue lengths than TODSum.  Meeting datasets, i.e., QMSum, and QA datasets, i.e., ConvoSumm, have a much longer length but these dialogues are mostly extracted from several special applications, thus not comparable to TODSum. Customer-service datasets like E-commerce are limited by privacy and not publicly available for research. Besides, we are the first to provide dialogue state knowledge to enhance the faithfulness and controllability of generated summaries.

\subsection{Evaluation Metrics}
% As a summarization dataset, ROUGE is still the main metric. In addition, we define a factual consistency evaluation method that are applicable to multi-domain task-based dialogue summarization datasets. We compare the predicted summary with all the slot values in the gold dialogue state. It is worth noting that each slot value needs the corresponding domain, intent and slot to be predicted correctly to be consistent with the facts. We follow this idea to calculate the values of fact-aware precision, recall and F1 score.

% We use two types of summarization metrics, ROUGE \cite{Lin2004AutomaticEO} and state-aware factual scores.
Along with the widely used automatic metric ROUGE scores \cite{Lin2004AutomaticEO}, we pay more attention to the factual consistency of generated summaries. For example, a user books a cheap hotel with 2 people but the model summarizes an expensive hotel with 3 people, which is a fatal mistake for practical applications. We use $N_{t}$ and $N_{h}$ to denote the number of \emph{(domain, intent, slot, value)} tuple in the target (golden dialogue state) and hypothesis (generated summary) \footnote{We extract related slot values from generated summaries based on the task ontology.} respectively. $\mathcal{N}(h \cap t)$ denotes the number of slot tuples exactly matched between generated summary and golden dialogue state. We define state-aware factual consistency metrics as follows: $\operatorname{Precision}=\mathcal{N}(h \cap t) / \mathcal{N}(h)$, $\operatorname{Recall}=\mathcal{N}(h \cap t) / \mathcal{N}(t)$. The former and latter respectively denote the percentage of exactly matched slot tuples in the generated summaries and golden dialogue states. We compute the overall F1 score using $\operatorname{F1}=2\cdot\operatorname{Precision}\cdot\operatorname{Recall}/(\operatorname{Precision}+\operatorname{Recall})$.

\section{State-Aware Structured Dialogue Summarization Model}
% lulu

\begin{figure}[t]
\centering{
\includegraphics[scale=0.65]{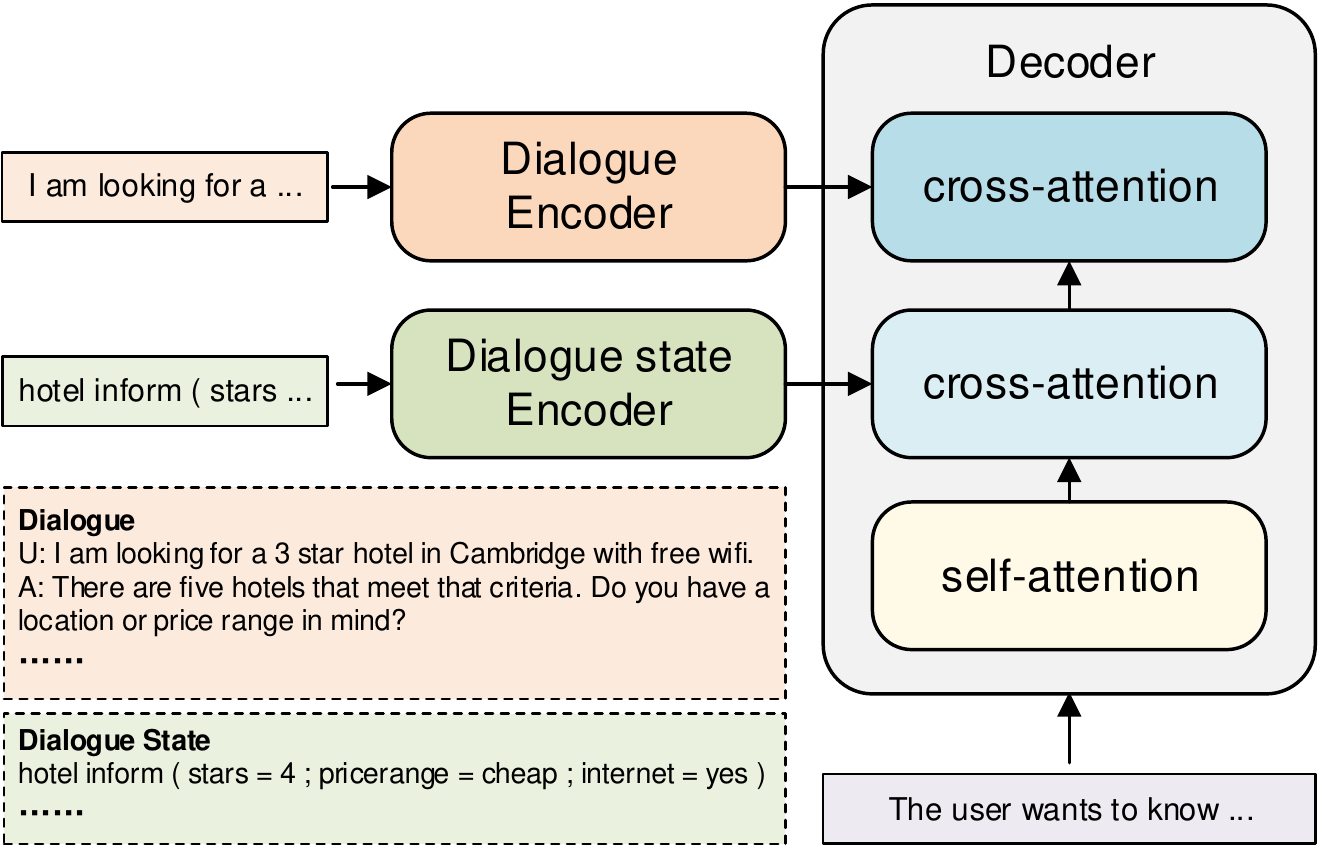}}
\caption{Overall architecture of our model.}
\label{fig:model}
\vspace{-0.6cm}
\end{figure}

% \usepackage{multirow}
\begin{table*}[]
\centering
\resizebox{1.0\textwidth}{!}{%
\begin{tabular}{l|l|ccc|ccc|ccc|ccc}
\hline
\multicolumn{2}{c|}{\multirow{2}{*}{\textbf{Model}}} & \multicolumn{3}{c|}{\textbf{ROUGE-1}} & \multicolumn{3}{c|}{\textbf{ROUGE-2}} & \multicolumn{3}{c|}{\textbf{ROUGE-L}} & \multicolumn{3}{c}{\textbf{Factuality}} \\ \cline{3-14} 
\multicolumn{2}{c|}{}                       & \textbf{F}        & \textbf{P}       & \textbf{R}       & \textbf{F}        & \textbf{P}       & \textbf{R}       & 
\textbf{F}        & \textbf{P}       & \textbf{R}       & 
\textbf{F}            & \textbf{P}           & \textbf{R}           \\ \hline
\multicolumn{1}{c|}{\multirow{3}{*}{Ext}}  &
Lead-3                                      & 24.06    & 27.99   & 22.43   & 5.01     & 6.29    & 4.52    & 20.27    & 23.63   & 18.85   & 21.74        & 29.53       & 17.20       \\
 & BertExt                                    & 41.49    & 42.87   & 42.10   & 13.13    & 14.94   & 12.45   & 36.11    & 37.42   & 36.55   & 21.61        & 21.88       & 21.35       \\
 & Oracle                                      & 46.07    & 49.04   & 44.90   & 15.66    & 18.41   & 14.24   & 39.79    & 42.45   & 38.69   & 23.11        & 24.21       & 22.10       \\
 \hline \hline
 \multicolumn{1}{c|}{\multirow{6}{*}{Abs}} 
 & BertAbs                                        & 71.37    & 67.86   & 78.11   & 54.12    & 52.52   & 59.26   & 68.89    & 65.53   & 75.36   & 49.45       & 49.39       & 49.51       \\
& BertAbs w. DS(pred) $\dagger$                            & 72.00    & 69.04   & 78.25   & 54.65    & 53.14   & 59.91   & 69.74    & 66.87   & 75.82   & 50.16        & 53.35       & 47.33       \\
& BertAbs w. DS(oracle) $\dagger$                          & \textbf{73.71}    & \textbf{70.25}   & \textbf{80.39}   & \textbf{57.11}    & \textbf{55.25}   & \textbf{62.60}   & \textbf{71.58}    & \textbf{68.23}   & \textbf{78.07}   & \textbf{53.66} & \textbf{57.07} & \textbf{50.63}       \\  \cline{2-14} 
& BART                                        & 70.90    & 68.10   & 76.14   & 55.65    & 53.39   & 60.62   & 68.23    & 65.57   & 73.21   & 55.53        & 56.66       & 54.45       \\
& BART w. DS(pred) $\dagger$                            & 72.46    & 79.67   & 72.46   & 57.82    & 54.02   & 64.91   & 69.71    & 65.61   & 76.59   & 61.21        & 61.68       & 60.75       \\
& BART w. DS(oracle) $\dagger$                         & \textbf{73.96}    & \textbf{69.81}   & \textbf{81.12}   & \textbf{60.66}    & \textbf{56.93}   & \textbf{67.87}   & \textbf{72.02}    & \textbf{68.00}   & \textbf{78.94}   & \textbf{66.15}        & \textbf{67.04}       & \textbf{65.28}       \\ \hline
\end{tabular}%
}
\vspace{-0.2cm}
\caption{ROUGE scores and factual consistency scores of different models on TODSum. Results are averaged over three random runs. $\dagger$ means our methods. DS(pred) and DS(oracle) denote automatic prediction and oracle extraction at test time respectively. ($p < 0.05$ under t-test)}
\label{tab:main-result}
\vspace{-0.6cm}
\end{table*}

\subsection{Model Architecture}
Figure \ref{fig:model} provides a description of our model. Inspired by \citet{zhu-etal-2021-enhancing} and \citet{dou-etal-2021-gsum}, we feed both the source dialogues and dialogue state information into the model. Note that it is a general framework based on the pre-trained models, i.e., BART and BERT, with the Transformer model \cite{NIPS2017_3f5ee243} as the backbone structure. To encode the dialogue state structures, we design a novel dialogue state encoder to complement the dialogue encoder.

For dialogue state encoder, the dialogue state information is concatenated as the input sequence : $input$$=$$domain_1$\ $intent_1(slot_1$$=$$value_1$\ $;slot_2$$=$$value_2)$\ $intent_2 (slot_1$$=$$value_1$\ $;slot_2$$=$$value_2)...$
The dialogue state encoder consists of multiple layers, each of which containing a self-attention block and a feed-forward block, just like the dialogue encoder. These two encoders share the parameters of the bottom layers and the word embedding layers, and only use different parameters in the last layer. Besides, in order to deal with the dialogue state information, we improve the standard Transformer. For each decoder layer, an additional cross-attention layer is added before the cross-attention layer of the dialogue to focus on the extraction of dialogue state information. In this case, the decoder will first attend to the dialogue state information and generate the corresponding representations, and then it will attend to the entire dialogue based on the dialogue state-aware information.

\subsection{Choices of Dialogue Structure}
% 这里强调下我们的加噪策略 model vs oracle
Following \citet{dou-etal-2021-gsum}, we feed two different variants of the dialogue state into our proposed model: 
\textbf{oracle extraction} and \textbf{automatic prediction}, called DS(oracle) and DS(pred) in subsequent experiments respectively.
As for oracle extraction, we use the golden dialogue state corresponding to the last turn of the dialogue. For automatic prediction, we use the dialogue state predicted by the dialogue state tracking model TRADE \cite{wu2019transferable}. It is worth noting that the introduction of these two dialogue structures does not require human participation. Subsequent experiments show that oracle extraction improves the model performance more than automatic prediction.

\subsection{Joint Dialogue Understanding and Summarization}
% 这里简要介绍，不做重点突出（实验效果并不好）
A simple joint model is also designed to verify the ability of our model to simultaneously generate the dialogue states and summaries. We modify the input sequence of the decoder of the above model: $input\_decoder$$=$$dialogue\ state$ \ $<$$|endoftext|$$>$\ $golden\ summary$. During training, the entire generated sequence is regarded as a whole to be calculated to get the final loss. However, different evaluation metrics are used to evaluate the dialogue states and summaries respectively respectively at test time.

\section{Experiments}

\subsection{Baselines}
We compare our methods with several baselines. The extractive baselines are included: (1) Oracle; (2) Lead-3; (3) BertExt \cite{liu-lapata-2019-text}. We also add pre-trained model-based abstractive methods for comparison: (1) BertAbs \cite{liu-lapata-2019-text}; (2) BART \cite{Lewis2020BARTDS}.\footnote{We will release our code at ***.} We give the details in Appendix.
% lulu

\subsection{Main Results}
% TODO： bart vs bert
% keqing
% 1. effect of dst: 加入DST信息后整体提升:
%    bert: 1:+2.34; 2:+2.99; l:+2.69; fact: +4.21
%    bart: 1:+3.06; 2:+5.01; l:+3.79; fact: +10.62
% 2. DST model vs oracle
%    bert:  1:+1.71, 2:+2.46, l:+1.84, fact:+3.50 
%    bart:  1:+1.50, 2:+2.84, l:+2.31, fact:+4.94 
% 3. 事实一致性指标
%    bert: unknown
%    bart: +10.62
Table \ref{tab:main-result} displays the main results of different models on TODSum. We perform experiments based on two strong baselines, BertAbs \cite{liu-lapata-2019-text} and BART \cite{Lewis2020BARTDS}. We also add several extractive summarization models for comparison.\footnote{Since key information scatters sparsely across the whole conversation, extractive models are extremely worse than abstractive ones. We only add dialogue state knowledge based on abstractive baselines.} We find that using the structured dialogue state information consistently outperforms both baselines on ROUGE and factual metrics. Specifically, BertAbs w. DS(oracle) outperforms BertAbs by 2.34\% on ROUGE-1, 2.99\% on ROUGE-2, and 2.69\% on ROUGE-L. BART w. DS(oracle) outperforms BART by 3.06\% on ROUGE-1, 5.01\% on ROUGE-2, and 3.79\% on ROUGE-L. The significant improvements prove the effectiveness of explicitly modeling dialogue states. For factual consistency metric F1, BertAbs w. DS(oracle) outperforms BertAbs by 4.21\% and BART w. DS(oracle) outperforms BART by 10.62\%. The higher improvements than ROUGE scores demonstrate our proposed state-aware structured dialogue summarization model significantly improves the faithfulness of generated summaries. It confirms that understanding dialogue structure is essential to factual consistency. Besides, comparing DS(pred) with DS(oracle), BART-based oracle outperforms pred by 1.50\% on ROUGE-1, 2.84\% on ROUGE-2, 2.31\% on ROUGE-L, which shows the quality of dialogue state information has an effect on the summarization performance. We leave the detailed analysis in the following section: Robustness.

\section{Qualitative Analysis}

\subsection{Dialogue Structure-Based Factual Consistency}
% fujia keqing
% TODO: 这里除了要分析指标外，还要对单独的错误类型进行归类分析，进而探讨dst为什么可以提升事实一致性
\begin{figure}[t]
\centering{
\includegraphics[scale=0.5]{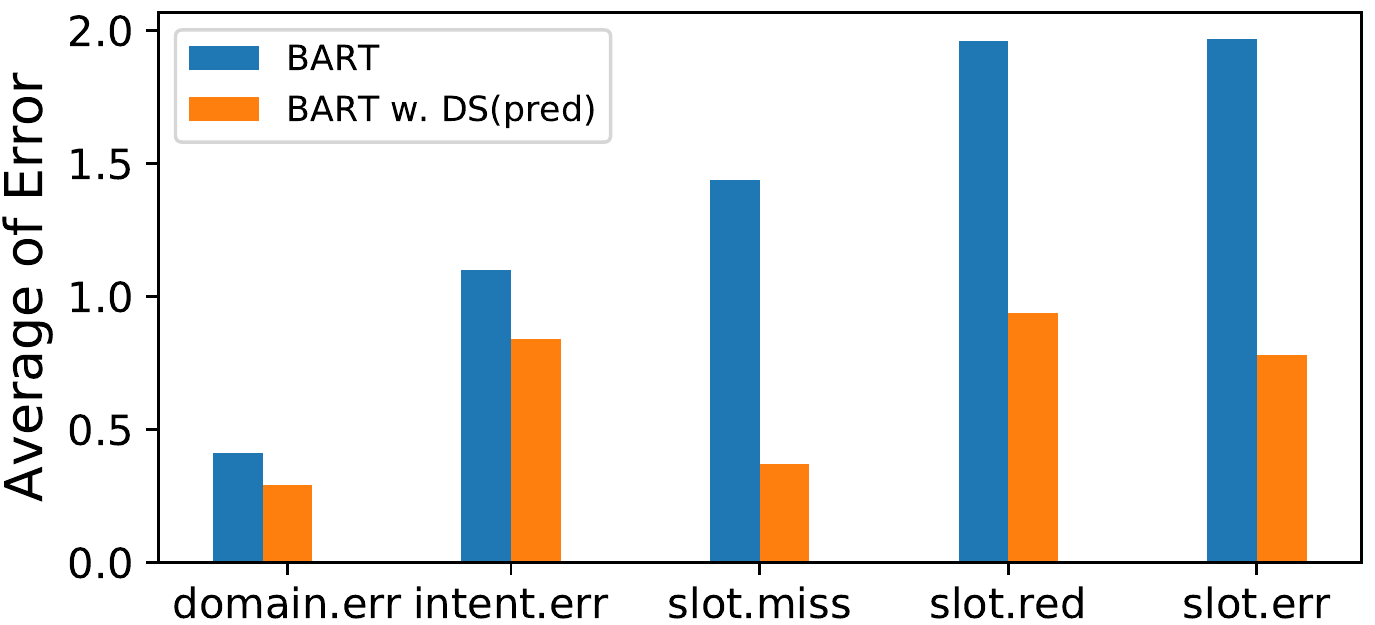}}
\vspace{-0.2cm}
\caption{Analysis of different types of factual errors.}
\label{fig:fac}
\vspace{-0.4cm}
\end{figure}

Fig \ref{fig:fac} displays the different types of factual errors of BART and BART w. DS(pred). We annotate the same set of 100 randomly sampled summaries and classify the errors into five types: \emph{domain error}, \emph{intent error}, \emph{slot missing}, \emph{slot redundancy}, and \emph{slot value error}. We report the average number of each error type that occurs on each sample and compare the difference between the two models. Results show that the BART baseline suffers from severe slot-related errors while our proposed BART w. DS can effectively alleviate these errors using dialogue state knowledge.

\subsection{Robustness Analysis}
\label{robustness}
% fujia keqing

\begin{figure}[t]
    \centering
    \subfigure[ROUGE scores]{
        \includegraphics[scale=0.24]{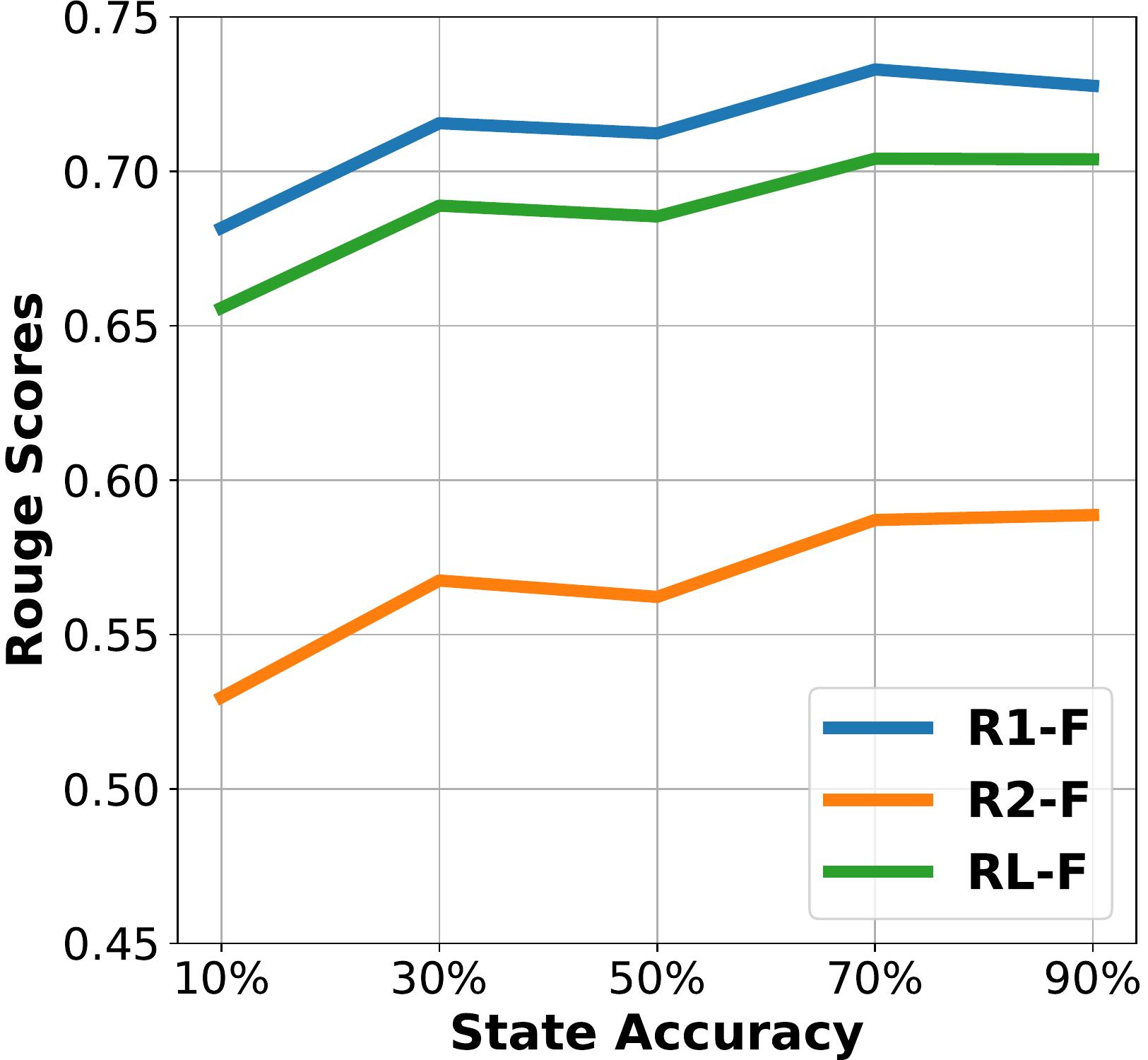}
    }
    \subfigure[Factual scores]{
        \includegraphics[scale=0.24]{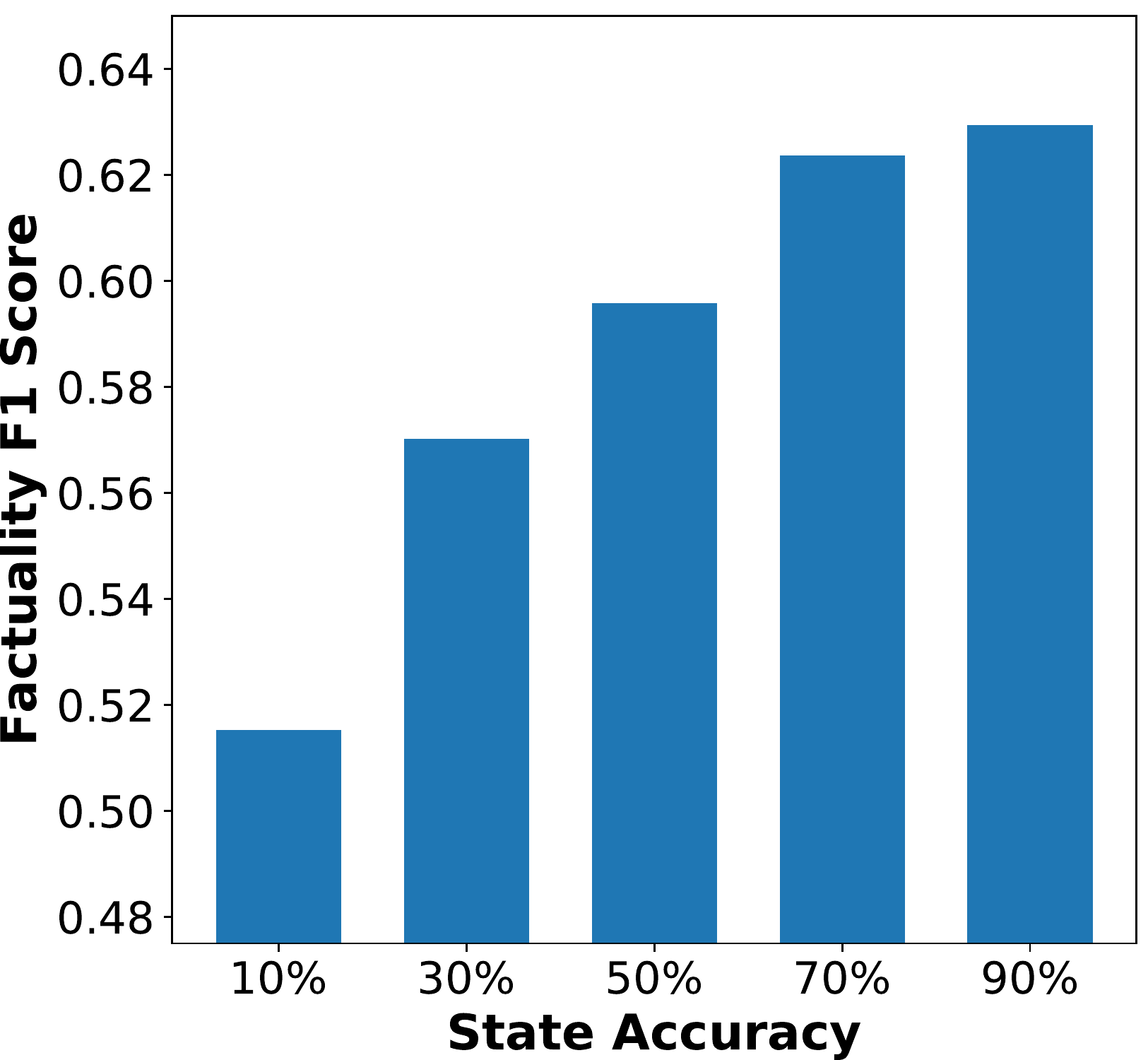}
    }
    \vspace{-0.3cm}
    \caption{Effect of different quality of dialogue states for training. The X-ray denotes the tuple-level accuracy of the noisy dialogue states.}
    \label{robust_train}
    \vspace{-0.3cm}
\end{figure}

\begin{figure}[t]
    \centering
    \subfigure[ROUGE scores]{
        \includegraphics[scale=0.24]{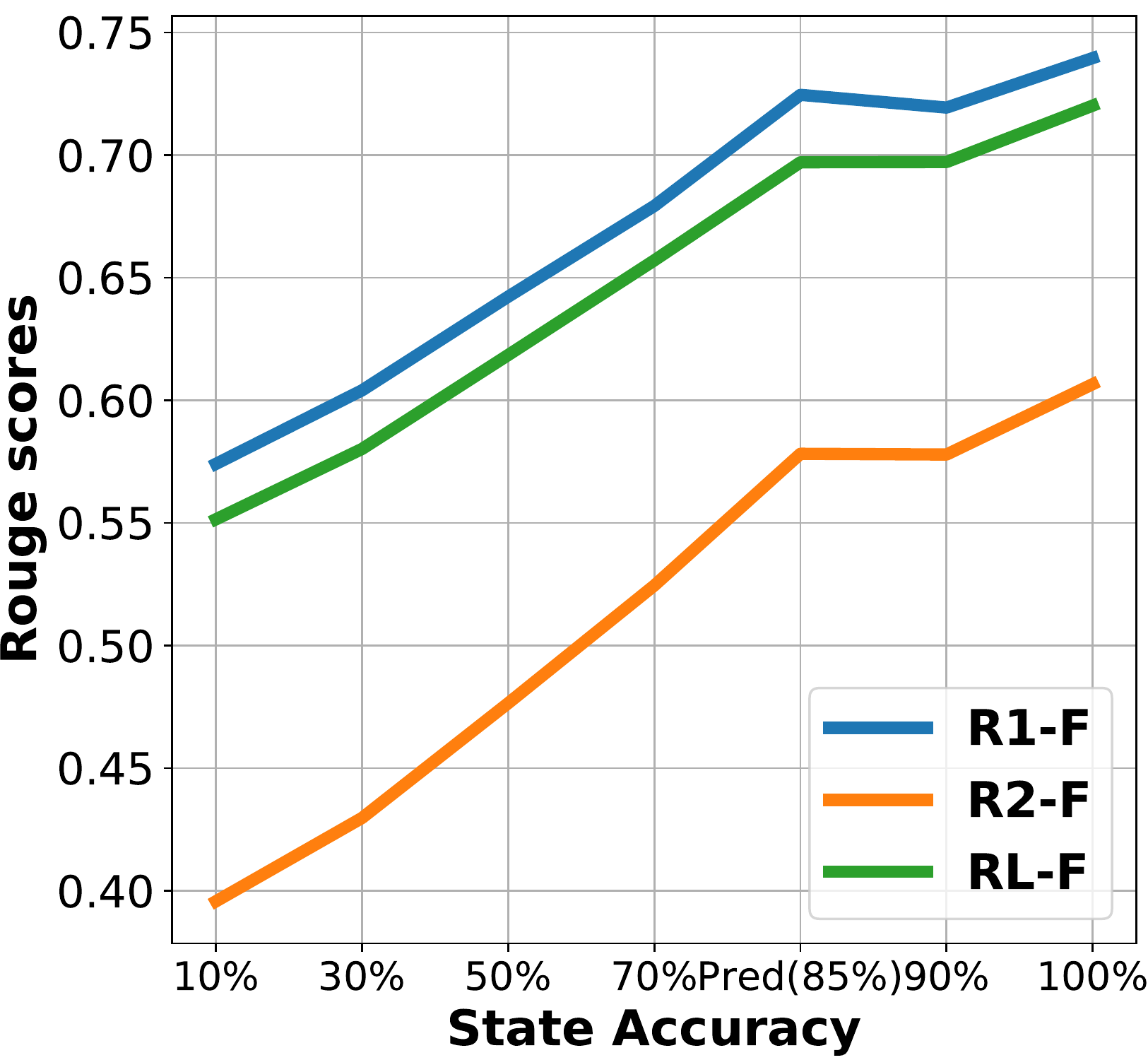}
    }
    \subfigure[Factual scores]{
        \includegraphics[scale=0.24]{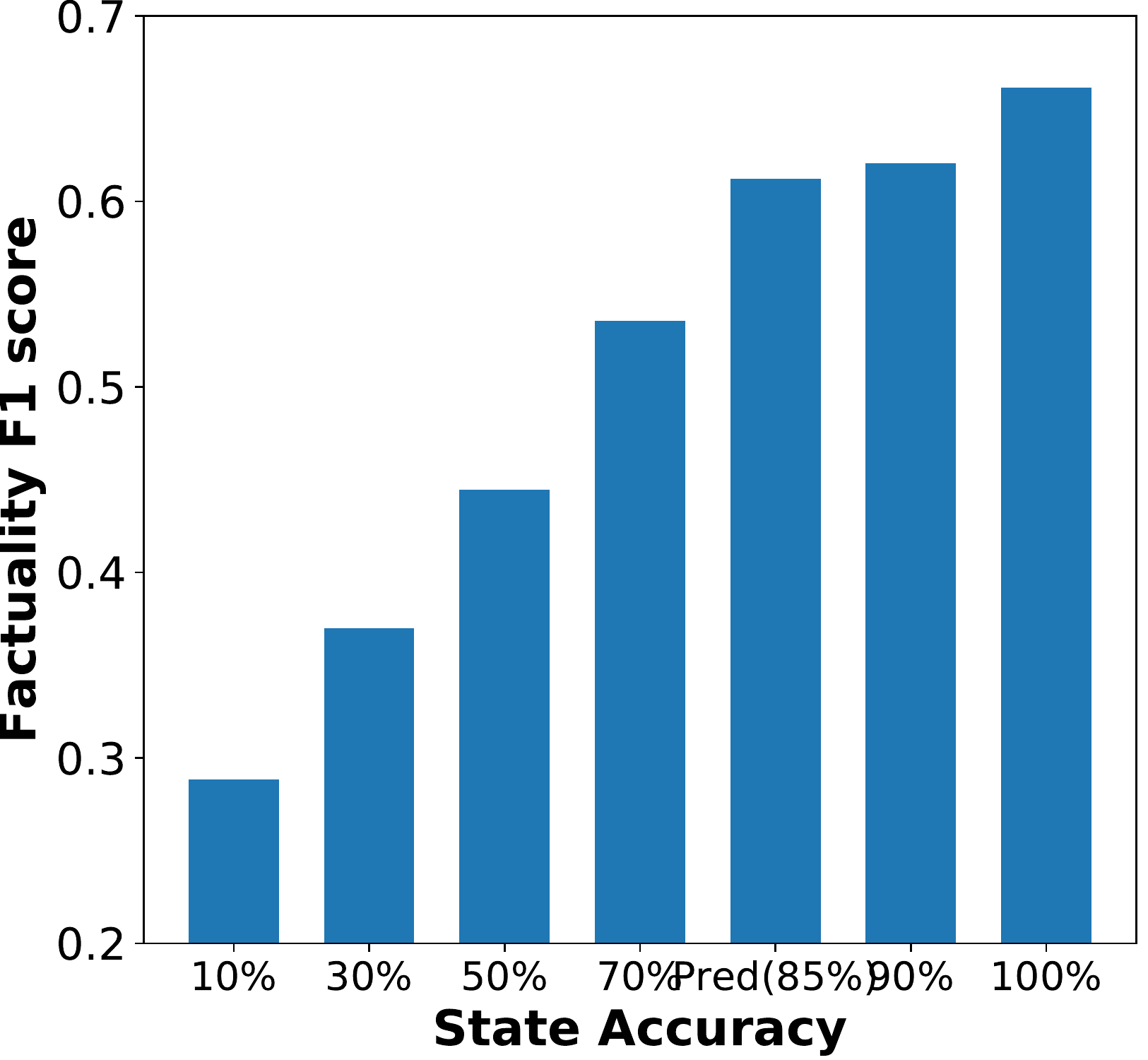}
    }
    \vspace{-0.4cm}
    \caption{Effect of different quality of dialogue states for test.}
    \label{robust_test}
    \vspace{-0.5cm}
\end{figure}

To verify the effect of the quality of dialogue state information, we perform the robustness analysis in this section. We construct the noisy variant dataset, Noisy-TODSum. Specifically, we add random noise to the oracle dialogue states, including delete \emph{(slot, value)} tuples, replace slot values, and insert new \emph{(slot, value)} tuples. We respectively add noise to the training set and the test set to show how the noisy state information affects the performance of summarization.

% 扰动加到训练集
\noindent\textbf{Training Robustness} Fig \ref{robust_train} shows the ROUGE and factual scores of our proposed BART w. DS(pred) under different noisy state information in the training set. We use the same oracle state for test. We find that, with the increase of state accuracy, BART w. DS(pred) gets better performance both on ROUGE and factual scores. Besides, the performance is similar when state accuracy is more than 70\%, which indicates we don't need to rely heavily on high-quality human annotated states and directly use existing DST models\footnote{For example, TRADE \cite{wu2019transferable} has already achieved almost 85\% tuple-level accuracy on MultiWOZ.}.

% 扰动加到测试集
\noindent\textbf{Test Robustness} Similarly, we also show the performance comparison (Fig \ref{robust_test}) under different noisy state information in the test set. Note that we add two special test accuracy, pred(acc=85\%) and oracle(acc=100\%) corresponding with BART w. DS(pred) and BART w. DS(oracle) in Table \ref{tab:main-result}. We find that higher test state accuracy results in better summarization performance. Comparing training noise with test noise, our model is more sensitive to the test noise, which proves dialogue state knowledge can effectively improve the controllability of generated summaries.

\subsection{Domain Adaptation}
\begin{figure}[t]
\centering{
\includegraphics[scale=0.5]{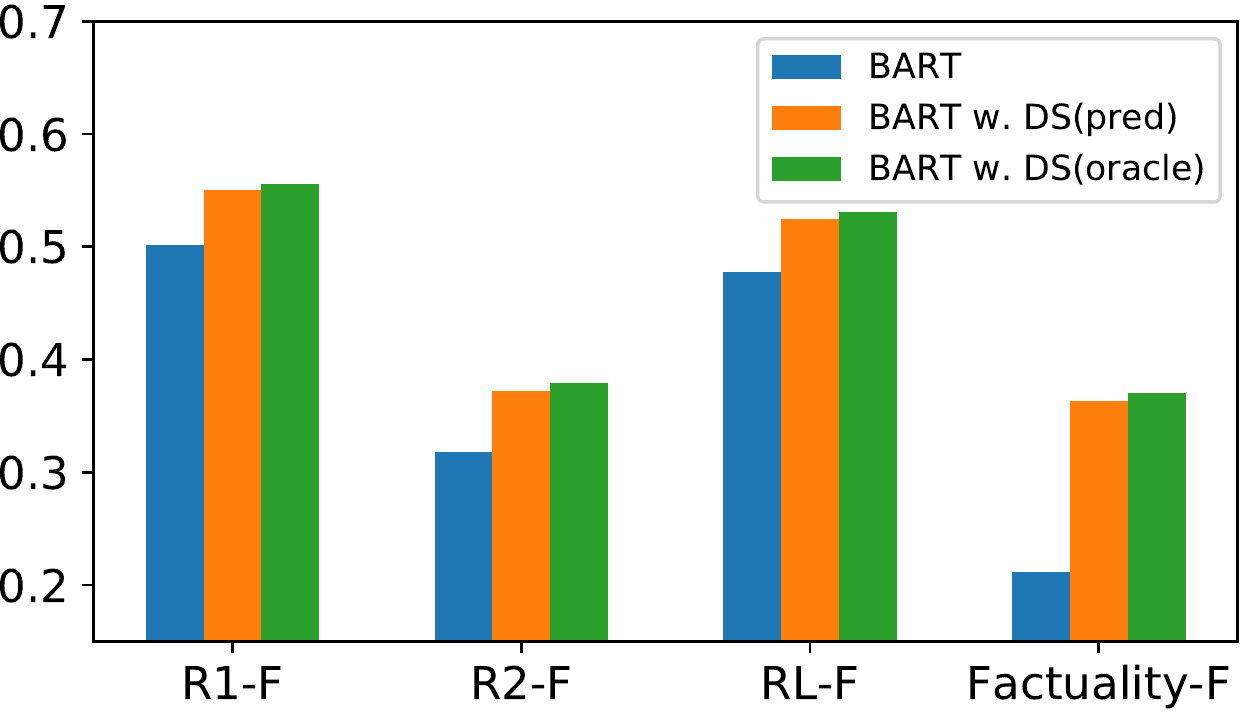}}
\vspace{-0.3cm}
\caption{Results of domain adaptation.}
\label{fig:domain}
\vspace{-0.3cm}
\end{figure}

% 领域迁移
% keqing
Fig \ref{fig:domain} shows the results of domain adaptation where we split dialogues into single domains, and choose four domains as training data, the other domain as test data, named DA-TODSum. We find zero-shot adaptation can't generate reasonable summaries thus we also sample 10\% target domain data for training to simulate few-shot domain adaptation. We choose each domain as the target domain and report the average scores for five runs. Results show our model with DS significantly outperforms BART, even higher than improvements of main results, which proves the effectiveness of dialogue state knowledge in the few-shot learning.

\subsection{Ablation Study}
% 1. 单纯的context vs dst，rouge得分持平，而dst事实性占优。说明dst对事实性有促进，但缺乏一些流畅度
% 2. 二者结合更优
\begin{table}[]
\centering
\small
\resizebox{0.45\textwidth}{!}{%
\begin{tabular}{l|cccc}
\hline
\textbf{Model}               & \textbf{R-1}   & \textbf{R-2}   & \textbf{R-L}   & \textbf{Factuality}\\ \hline
BART                & 70.90 & 55.65 & 68.23  & 55.53\\
BART only DS(pred)   & 71.03 & 55.28 & 67.95  & 60.92\\
BART w. DS(pred)     & \textbf{72.46} & \textbf{57.82} & \textbf{69.71}  & \textbf{61.21}\\ \hline
BART only DS(oracle) & 73.56 & 58.71 & 71.30  & 65.54\\
BART w. DS(oracle)   & \textbf{73.96} & \textbf{60.66} & \textbf{72.02}  & \textbf{66.15}\\
\hline
\end{tabular}%
}
\vspace{-0.1cm}
\caption{F1 scores of ablation study. BART denotes only using original dialogue history. Instead, BART only DS takes dialogue states as the only input.}
\label{tab:ablation}
\vspace{-0.6cm}
\end{table}

We display the ablation study in Table \ref{tab:ablation} to show the importance of different input sources. Comparing dialogue history with pred state, only using pred state gets similar ROUGE scores but much higher factual scores (+5.39\%), which proves dialogue state helps summarization models enhance faithfulness but get a drop in fluency. Combining both as input, our proposed state-aware structured dialogue summarization model gets the best performance for ROUGE and factual scores.

\subsection{Joint Model}
% keqing
Table \ref{tab:joint-model} shows the result of the joint model. We report the summarization metrics, ROUGE and factual scores, and dialogue state tracking metrics, Turn Acc. We find the joint model gets a worse performance than BART baseline on ROUGE and factual scores, and a lower acc than the DST model, TRADE(98.76). Although the joint model is simple and weak, more future work for joint dialogue understanding and summarization is worth being explored, like graph neural networks, better combination of state tracking task and summarization task, etc.
% \usepackage{multirow}
\begin{table}[]
\small
\centering
\resizebox{0.48\textwidth}{!}{%
\begin{tabular}{l|ccccc}
\hline
\textbf{Model}    & \textbf{R-1} & \textbf{R-2} & \textbf{R-L} & \multicolumn{1}{l}{\textbf{Factuality}} & \textbf{Acc}\\ \hline
BART               & 70.90        & 55.65        & 68.23        & 55.53     & -                              \\
BART w. DS(pred)   & 72.46        & 57.82        & 69.71        & 61.21         & -                          \\                        
Joint Model        & 65.75        & 49.58        & 63.06        & 43.18         &       75.40                    \\ \hline
% \textbf{Models}    & \multicolumn{4}{c}{\textbf{Turn.Acc}}                                               \\ \hline
% TRADE              & \multicolumn{4}{c}{98.76}                                                           \\
% Joint Model        & \multicolumn{4}{c}{75.40}                                                           \\ \hline
\end{tabular}%
}
\vspace{-0.1cm}
\caption{Results of joint model. Acc denotes the turn-level accuracy of dialogue state tracking.}
\label{tab:joint-model}
\vspace{-0.6cm}
\end{table}

\subsection{Human Evaluation}
% lulu
We conduct a manual evaluation to assess the models. 100 samples are randomly selected from the TODSum and five native speakers of English are hired to rate the ground truth and summaries generated by different models. Each annotator scores summaries from 1 (worst) to 5 (best) on fluency, informativeness, redundancy, and factualness. Each instance is rated by three annotators and the scores for each summary are averaged. The intra-class agreement score is 0.592.

As shown in Table \ref{tab:human_evaluation}, BertAbs w. DS(pred) and BART w. DS(pred) achieve great progress in the four metrics than BertAbs and BART respectively, especially in terms of factualness. This suggests the dialogue state information enhances the ability of models to identify the key information, such as slots and intents. However, for two pre-trained models, BertAbs w. DS(pred) performs worse than BART w. DS(pred) in factualness, which demonstrates the randomly initialized decoder in BertAbs can not pay attention to the structured information in the dialogues. In addition, all models perform poorly in redundancy. This is because the model emphasizes informativeness to the greatest extent, and there is no adequate balance between it and redundancy.

\subsection{Case Study}
Fig 2 in Appendix shows two examples from the TODSum dataset. For Example one, BART incorrectly generated the intent of the user about train domain, that is, the id and arrival time of the train are incorrectly recognized as the number of people. Besides, the departure and destination of the train are reversed, which belongs to wrong generation of slot values. The most serious is that BART misunderstands the intent of the user about restaurants as the attention to the attraction, which leads to the incorrect subsequent field information. For Example two, the summary generated by BART loses some slots and corresponding values, such as the name of hotel. Additionally, all information for the taxi domain is missing, such as the type of the car, the driver's phone number, and the leaving time of the taxi. According to observations, our model avoids these errors to some extent.  By designing a novel dialogue state encoder, the dialogue state features are well represented, and a dialogue state-based cross-attention layer guides our model to focus on dialogue structures, which makes generated summaries identify more key information and contain more faithful facts. However, there are some deficiencies in redundancy for our models, such as hotels with parking and internet.

%补一个图，包含2个例子，一个是关于改进信息缺失的（意图、slot）；一个是关于错误推理的（领域、意图、slot）一段分析

% \subsection{Visualization}
% lulu

\section{Discussion}
\begin{table}[]
\centering
\small
\begin{tabular}{l|cccc}
\hline
\textbf{Model} & \textbf{Flu.} & \textbf{Inf.} & \textbf{Red.} & \textbf{Fac.} \\
\hline
Ground Truth & 4.93 & 4.58 & 4.37 & 4.26 \\
BertAbs & 4.20 & 4.13 & 3.51 & 3.19 \\
BertAbs w. DS(pred) & 4.37 & 4.29 & 3.57 & 3.28 \\
BART & 4.18 & 4.22 & 3.55 & 3.57 \\
BART w. DS(pred) & 4.41 & 4.36 & 3.63 & 4.12\\
\hline
\end{tabular}
\vspace{-0.1cm}
\caption{Human evaluation on Fluency (Flu.), Informativeness (Inf.), Redundancy (Red.), and Factualness (Fac.)}
\label{tab:human_evaluation}
\vspace{-0.6cm}
\end{table}

% lulu keqing
To further understand the current issues of TODSum, we perform error analysis on generated summaries. Then we provide several potential directions to handle these issues for future work.

\subsection{Error Analysis}
Compared to the golden summaries, we observe the generated summaries  by existing models and summarize the following three error types:

\textbf{1. Information Missing}: The salient elements mentioned in golden summaries, i.e., slot and intent, are missing in generated summaries.

\textbf{2. Lack of Reasoning} Current models lack context reasoning capability while facing multi-turn negotiation. They just memorize the surface words or phrases but disregard the logical relationships between conversations, leading to the wrong selection and combination of salient elements.

\textbf{3. Redundancy} The golden summaries do not mention what appears in the generated summaries, especially the domains and intents.

% We use the above error type taxonomy to annotate 100 examples randomly selected from TODSum. In this case, one summary might have more than one labels. If the summary do not belong to any error types, we will categorize it as (0) Other.
% 加一个错误类型的表格，一个分析

\subsection{Future Directions}
Here we discuss several potential future directions worth being explored. 
\begin{itemize}
    \item Graph Neural Network. Recent work \cite{Chen2020MultiViewSM, Chen2021StructureAwareAC} have been exploring graph-based summarization methods but are limited by noisy unsupervised signals like topic modeling, discourse graphs. These signals are coarse-grained and not intended for dialogues. Based on the dialogue state, we can better understand the dialogue context for summarization generation.
    \item Multi-Task Learning. Although the current joint dialogue understanding and summarization model doesn't achieve a good performance. We still believe understanding and summarization are both important for each other. Even, the design of a task-oriented dialogue system can be combined with a summarization module.
    \item Factual Consistency. The dialogue state provides a uniform and simple evaluation tool to compute factual consistency, which may help the progress of related work.
\end{itemize}
% In our opinion, there are three interesting directions for future work. The first is to combine the slots and graphs to structurely represent the dialogue state information. The second is to design a joint model that generates dialogue states and summaries at the same time based on the multi-task learning. The third is to ensure the redundancy, informativeness, and factual consistency at the same time.
\section{Conclusion}
% keqing
In this paper, we introduce a task-oriented dialogue summarization dataset, TODSum. We hope to utilize the task-oriented dialogue structure to enhance the faithfulness and controllability of generated summaries.
Thus we propose a general dialogue state-aware summarization model to combine original dialogue text and structured dialogue state. We conduct exhaustive qualitative analysis to show the effectiveness of dialogue structure guidance and discuss current challenges of TODSum and potential development directions for future work.

% \section{Acknowledgments}

\bibliography{aaai22.bib}

\begin{thebibliography}{46}
\providecommand{\natexlab}[1]{#1}

\bibitem[{Budzianowski et~al.(2018)Budzianowski, Wen, Tseng, Casanueva, Ultes,
  Ramadan, and Ga{\v{s}}i{\'c}}]{budzianowski2018multiwoz}
Budzianowski, P.; Wen, T.-H.; Tseng, B.-H.; Casanueva, I.; Ultes, S.; Ramadan,
  O.; and Ga{\v{s}}i{\'c}, M. 2018.
\newblock MultiWOZ--A Large-Scale Multi-Domain Wizard-of-Oz Dataset for
  Task-Oriented Dialogue Modelling.
\newblock \emph{arXiv preprint arXiv:1810.00278}.

\bibitem[{Cao et~al.(2020)Cao, Dong, Wu, and Cheung}]{cao-etal-2020-factual}
Cao, M.; Dong, Y.; Wu, J.; and Cheung, J. C.~K. 2020.
\newblock Factual Error Correction for Abstractive Summarization Models.
\newblock In \emph{Proceedings of the 2020 Conference on Empirical Methods in
  Natural Language Processing (EMNLP)}, 6251--6258. Online: Association for
  Computational Linguistics.

\bibitem[{Cao et~al.(2017)Cao, Wei, Li, and Li}]{cao2017faithful}
Cao, Z.; Wei, F.; Li, W.; and Li, S. 2017.
\newblock Faithful to the Original: Fact Aware Neural Abstractive
  Summarization.
\newblock In \emph{Proceedings of the 32th AAAI Conference on Artificial
  Intelligence}.

\bibitem[{Carletta et~al.(2005)Carletta, Ashby, Bourban, Flynn, Guillemot,
  Hain, Kadlec, Karaiskos, Kraaij, Kronenthal, Lathoud, Lincoln, Masson,
  McCowan, Post, Reidsma, and Wellner}]{Carletta2005TheAM}
Carletta, J.; Ashby, S.; Bourban, S.; Flynn, M.; Guillemot, M.; Hain, T.;
  Kadlec, J.; Karaiskos, V.; Kraaij, W.; Kronenthal, M.; Lathoud, G.; Lincoln,
  M.; Masson, A.~L.; McCowan, I.; Post, W.; Reidsma, D.; and Wellner, P. 2005.
\newblock The AMI Meeting Corpus: A Pre-announcement.
\newblock In \emph{MLMI}.

\bibitem[{Chen and Yang(2020)}]{Chen2020MultiViewSM}
Chen, J.; and Yang, D. 2020.
\newblock Multi-View Sequence-to-Sequence Models with Conversational Structure
  for Abstractive Dialogue Summarization.
\newblock In \emph{EMNLP}.

\bibitem[{Chen and Yang(2021)}]{Chen2021StructureAwareAC}
Chen, J.; and Yang, D. 2021.
\newblock Structure-Aware Abstractive Conversation Summarization via Discourse
  and Action Graphs.
\newblock In \emph{NAACL}.

\bibitem[{Chen et~al.(2021)Chen, Liu, Chen, and Zhang}]{Chen2021DialogSumAR}
Chen, Y.; Liu, Y.; Chen, L.; and Zhang, Y. 2021.
\newblock DialogSum: A Real-Life Scenario Dialogue Summarization Dataset.
\newblock In \emph{FINDINGS}.

\bibitem[{Chowdhury and Chakraborty(2019)}]{Chowdhury2019CQASUMMBR}
Chowdhury, T.; and Chakraborty, T. 2019.
\newblock CQASUMM: Building References for Community Question Answering
  Summarization Corpora.
\newblock \emph{Proceedings of the ACM India Joint International Conference on
  Data Science and Management of Data}.

\bibitem[{Devlin et~al.(2019)Devlin, Chang, Lee, and
  Toutanova}]{Devlin2019BERTPO}
Devlin, J.; Chang, M.-W.; Lee, K.; and Toutanova, K. 2019.
\newblock BERT: Pre-training of Deep Bidirectional Transformers for Language
  Understanding.
\newblock In \emph{NAACL}.

\bibitem[{Dong et~al.(2020)Dong, Wang, Gan, Cheng, Cheung, and
  Liu}]{dong-etal-2020-multi}
Dong, Y.; Wang, S.; Gan, Z.; Cheng, Y.; Cheung, J. C.~K.; and Liu, J. 2020.
\newblock Multi-Fact Correction in Abstractive Text Summarization.
\newblock In \emph{Proceedings of the 2020 Conference on Empirical Methods in
  Natural Language Processing (EMNLP)}, 9320--9331. Online: Association for
  Computational Linguistics.

\bibitem[{Dou et~al.(2021)Dou, Liu, Hayashi, Jiang, and
  Neubig}]{dou-etal-2021-gsum}
Dou, Z.-Y.; Liu, P.; Hayashi, H.; Jiang, Z.; and Neubig, G. 2021.
\newblock {GS}um: A General Framework for Guided Neural Abstractive
  Summarization.
\newblock In \emph{Proceedings of the 2021 Conference of the North American
  Chapter of the Association for Computational Linguistics: Human Language
  Technologies}, 4830--4842. Online: Association for Computational Linguistics.

\bibitem[{Durmus, He, and Diab(2020)}]{durmus-etal-2020-feqa}
Durmus, E.; He, H.; and Diab, M. 2020.
\newblock {FEQA}: A Question Answering Evaluation Framework for Faithfulness
  Assessment in Abstractive Summarization.
\newblock In \emph{Proceedings of the 58th Annual Meeting of the Association
  for Computational Linguistics}, 5055--5070. Online: Association for
  Computational Linguistics.

\bibitem[{Fabbri et~al.(2021)Fabbri, Rahman, Rizvi, Wang, Li, Mehdad, and
  Radev}]{Fabbri2021ConvoSummCS}
Fabbri, A.~R.; Rahman, F.; Rizvi, I.; Wang, B.; Li, H.; Mehdad, Y.; and Radev,
  D. 2021.
\newblock ConvoSumm: Conversation Summarization Benchmark and Improved
  Abstractive Summarization with Argument Mining.
\newblock In \emph{ACL/IJCNLP}.

\bibitem[{Falke et~al.(2019)Falke, Ribeiro, Utama, Dagan, and
  Gurevych}]{falke-etal-2019-ranking}
Falke, T.; Ribeiro, L. F.~R.; Utama, P.~A.; Dagan, I.; and Gurevych, I. 2019.
\newblock Ranking Generated Summaries by Correctness: An Interesting but
  Challenging Application for Natural Language Inference.
\newblock In \emph{Proceedings of the 57th Annual Meeting of the Association
  for Computational Linguistics}, 2214--2220. Florence, Italy: Association for
  Computational Linguistics.

\bibitem[{Gliwa et~al.(2019)Gliwa, Mochol, Biesek, and
  Wawer}]{Gliwa2019SAMSumCA}
Gliwa, B.; Mochol, I.; Biesek, M.; and Wawer, A. 2019.
\newblock SAMSum Corpus: A Human-annotated Dialogue Dataset for Abstractive
  Summarization.
\newblock \emph{ArXiv}, abs/1911.12237.

\bibitem[{Goodrich et~al.(2019)Goodrich, Rao, Liu, and
  Saleh}]{10.1145/3292500.3330955}
Goodrich, B.; Rao, V.; Liu, P.~J.; and Saleh, M. 2019.
\newblock Assessing The Factual Accuracy of Generated Text.
\newblock In \emph{Proceedings of the 25th ACM SIGKDD International Conference
  on Knowledge Discovery \& Data Mining}, KDD19, 166–175. New York, NY, USA:
  Association for Computing Machinery.
\newblock ISBN 9781450362016.

\bibitem[{He, Yan, and Xu(2020)}]{he-etal-2020-learning-tag}
He, K.; Yan, Y.; and Xu, W. 2020.
\newblock Learning to Tag {OOV} Tokens by Integrating Contextual Representation
  and Background Knowledge.
\newblock In \emph{Proceedings of the 58th Annual Meeting of the Association
  for Computational Linguistics}, 619--624. Online: Association for
  Computational Linguistics.

\bibitem[{Huang, Zhu, and Gao(2020)}]{Huang2020ChallengesIB}
Huang, M.; Zhu, X.; and Gao, J. 2020.
\newblock Challenges in Building Intelligent Open-domain Dialog Systems.
\newblock \emph{ACM Transactions on Information Systems (TOIS)}, 38: 1 -- 32.

\bibitem[{Janin et~al.(2003)Janin, Baron, Edwards, Ellis, Gelbart, Morgan,
  Peskin, Pfau, Shriberg, Stolcke, and Wooters}]{1198793}
Janin, A.; Baron, D.; Edwards, J.; Ellis, D.; Gelbart, D.; Morgan, N.; Peskin,
  B.; Pfau, T.; Shriberg, E.; Stolcke, A.; and Wooters, C. 2003.
\newblock The ICSI Meeting Corpus.
\newblock In \emph{2003 IEEE International Conference on Acoustics, Speech, and
  Signal Processing, 2003. Proceedings. (ICASSP '03).}, volume~1, I--I.

\bibitem[{Kedzie, McKeown, and Daum{\'e}(2018)}]{Kedzie2018ContentSI}
Kedzie, C.; McKeown, K.; and Daum{\'e}, H. 2018.
\newblock Content Selection in Deep Learning Models of Summarization.
\newblock In \emph{EMNLP}.

\bibitem[{Koncel-Kedziorski et~al.(2019)Koncel-Kedziorski, Bekal, Luan, Lapata,
  and Hajishirzi}]{KoncelKedziorski2019TextGF}
Koncel-Kedziorski, R.; Bekal, D.; Luan, Y.; Lapata, M.; and Hajishirzi, H.
  2019.
\newblock Text Generation from Knowledge Graphs with Graph Transformers.
\newblock In \emph{NAACL}.

\bibitem[{Kryscinski et~al.(2020)Kryscinski, McCann, Xiong, and
  Socher}]{kryscinski-etal-2020-evaluating}
Kryscinski, W.; McCann, B.; Xiong, C.; and Socher, R. 2020.
\newblock Evaluating the Factual Consistency of Abstractive Text Summarization.
\newblock In \emph{Proceedings of the 2020 Conference on Empirical Methods in
  Natural Language Processing (EMNLP)}, 9332--9346. Online: Association for
  Computational Linguistics.

\bibitem[{Lewis et~al.(2020)Lewis, Liu, Goyal, Ghazvininejad, Mohamed, Levy,
  Stoyanov, and Zettlemoyer}]{Lewis2020BARTDS}
Lewis, M.; Liu, Y.; Goyal, N.; Ghazvininejad, M.; Mohamed, A.; Levy, O.;
  Stoyanov, V.; and Zettlemoyer, L. 2020.
\newblock BART: Denoising Sequence-to-Sequence Pre-training for Natural
  Language Generation, Translation, and Comprehension.
\newblock In \emph{ACL}.

\bibitem[{Li et~al.(2018)Li, Zhu, Zhang, and Zong}]{li-etal-2018-ensure}
Li, H.; Zhu, J.; Zhang, J.; and Zong, C. 2018.
\newblock Ensure the Correctness of the Summary: Incorporate Entailment
  Knowledge into Abstractive Sentence Summarization.
\newblock In \emph{Proceedings of the 27th International Conference on
  Computational Linguistics}, 1430--1441. Santa Fe, New Mexico, USA:
  Association for Computational Linguistics.

\bibitem[{Lin and Och(2004)}]{Lin2004AutomaticEO}
Lin, C.-Y.; and Och, F. 2004.
\newblock Automatic Evaluation of Machine Translation Quality Using Longest
  Common Subsequence and Skip-Bigram Statistics.
\newblock In \emph{ACL}.

\bibitem[{Liu et~al.(2019)Liu, Wang, Xu, Li, and Ye}]{Liu2019AutomaticDS}
Liu, C.; Wang, P.; Xu, J.; Li, Z.; and Ye, J. 2019.
\newblock Automatic Dialogue Summary Generation for Customer Service.
\newblock \emph{Proceedings of the 25th ACM SIGKDD International Conference on
  Knowledge Discovery \& Data Mining}.

\bibitem[{Liu and Lapata(2019)}]{liu-lapata-2019-text}
Liu, Y.; and Lapata, M. 2019.
\newblock Text Summarization with Pretrained Encoders.
\newblock In \emph{Proceedings of the 2019 Conference on Empirical Methods in
  Natural Language Processing and the 9th International Joint Conference on
  Natural Language Processing (EMNLP-IJCNLP)}, 3730--3740. Hong Kong, China:
  Association for Computational Linguistics.

\bibitem[{Nallapati et~al.(2016)Nallapati, Zhou, Santos, Çaglar G{\"u}lçehre,
  and Xiang}]{Nallapati2016AbstractiveTS}
Nallapati, R.; Zhou, B.; Santos, C.~D.; Çaglar G{\"u}lçehre; and Xiang, B.
  2016.
\newblock Abstractive Text Summarization using Sequence-to-sequence RNNs and
  Beyond.
\newblock In \emph{CoNLL}.

\bibitem[{Peng et~al.(2020)Peng, Zhu, Li, Li, Li, Zeng, and Gao}]{peng2020few}
Peng, B.; Zhu, C.; Li, C.; Li, X.; Li, J.; Zeng, M.; and Gao, J. 2020.
\newblock Few-shot natural language generation for task-oriented dialog.
\newblock \emph{arXiv preprint arXiv:2002.12328}.

\bibitem[{Radev, Hovy, and McKeown(2002)}]{Radev2002IntroductionTT}
Radev, D.~R.; Hovy, E.; and McKeown, K. 2002.
\newblock Introduction to the Special Issue on Summarization.
\newblock \emph{Computational Linguistics}, 28: 399--408.

\bibitem[{Sharma, Li, and Wang(2019)}]{Sharma2019BIGPATENTAL}
Sharma, E.; Li, C.; and Wang, L. 2019.
\newblock BIGPATENT: A Large-Scale Dataset for Abstractive and Coherent
  Summarization.
\newblock In \emph{ACL}.

\bibitem[{Vaswani et~al.(2017)Vaswani, Shazeer, Parmar, Uszkoreit, Jones,
  Gomez, Kaiser, and Polosukhin}]{NIPS2017_3f5ee243}
Vaswani, A.; Shazeer, N.; Parmar, N.; Uszkoreit, J.; Jones, L.; Gomez, A.~N.;
  Kaiser, L.~u.; and Polosukhin, I. 2017.
\newblock Attention is All you Need.
\newblock In Guyon, I.; Luxburg, U.~V.; Bengio, S.; Wallach, H.; Fergus, R.;
  Vishwanathan, S.; and Garnett, R., eds., \emph{Advances in Neural Information
  Processing Systems}, volume~30. Curran Associates, Inc.

\bibitem[{Wang, Cho, and Lewis(2020)}]{wang-etal-2020-asking}
Wang, A.; Cho, K.; and Lewis, M. 2020.
\newblock Asking and Answering Questions to Evaluate the Factual Consistency of
  Summaries.
\newblock In \emph{Proceedings of the 58th Annual Meeting of the Association
  for Computational Linguistics}, 5008--5020. Online: Association for
  Computational Linguistics.

\bibitem[{Wang et~al.(2020)Wang, Lin, Zhong, and Wong}]{Wang2020DialogueST}
Wang, D.; Lin, C.; Zhong, L.; and Wong, K.-F. 2020.
\newblock Dialogue State Tracking with Pretrained Encoder for Multi-domain
  Trask-oriented Dialogue Systems.
\newblock \emph{ArXiv}, abs/2004.10663.

\bibitem[{Wu et~al.(2019)Wu, Madotto, Hosseini-Asl, Xiong, Socher, and
  Fung}]{wu2019transferable}
Wu, C.-S.; Madotto, A.; Hosseini-Asl, E.; Xiong, C.; Socher, R.; and Fung, P.
  2019.
\newblock Transferable multi-domain state generator for task-oriented dialogue
  systems.
\newblock \emph{arXiv preprint arXiv:1905.08743}.

\bibitem[{Yan et~al.(2020)Yan, He, Xu, Liu, Meng, Hu, and
  Xu}]{yan-etal-2020-adversarial}
Yan, Y.; He, K.; Xu, H.; Liu, S.; Meng, F.; Hu, M.; and Xu, W. 2020.
\newblock Adversarial Semantic Decoupling for Recognizing Open-Vocabulary
  Slots.
\newblock In \emph{Proceedings of the 2020 Conference on Empirical Methods in
  Natural Language Processing (EMNLP)}, 6070--6075. Online: Association for
  Computational Linguistics.

\bibitem[{Yuan and Yu(2019)}]{Yuan2019AbstractiveDS}
Yuan, L.; and Yu, Z. 2019.
\newblock Abstractive Dialog Summarization with Semantic Scaffolds.
\newblock \emph{ArXiv}, abs/1910.00825.

\bibitem[{Zhang et~al.(2020{\natexlab{a}})Zhang, Zhao, Saleh, and
  Liu}]{Zhang2020PEGASUSPW}
Zhang, J.; Zhao, Y.; Saleh, M.; and Liu, P.~J. 2020{\natexlab{a}}.
\newblock PEGASUS: Pre-training with Extracted Gap-sentences for Abstractive
  Summarization.
\newblock \emph{ArXiv}, abs/1912.08777.

\bibitem[{Zhang et~al.(2020{\natexlab{b}})Zhang, Merck, Tsai, Manning, and
  Langlotz}]{zhang-etal-2020-optimizing}
Zhang, Y.; Merck, D.; Tsai, E.; Manning, C.~D.; and Langlotz, C.
  2020{\natexlab{b}}.
\newblock Optimizing the Factual Correctness of a Summary: A Study of
  Summarizing Radiology Reports.
\newblock In \emph{Proceedings of the 58th Annual Meeting of the Association
  for Computational Linguistics}, 5108--5120. Online: Association for
  Computational Linguistics.

\bibitem[{Zhang et~al.(2020{\natexlab{c}})Zhang, Takanobu, Huang, and
  Zhu}]{Zhang2020RecentAA}
Zhang, Z.; Takanobu, R.; Huang, M.; and Zhu, X. 2020{\natexlab{c}}.
\newblock Recent Advances and Challenges in Task-oriented Dialog System.
\newblock \emph{ArXiv}, abs/2003.07490.

\bibitem[{Zhong et~al.(2021)Zhong, Yin, Yu, Zaidi, Mutuma, Jha, Awadallah,
  Çelikyilmaz, Liu, Qiu, and Radev}]{Zhong2021QMSumAN}
Zhong, M.; Yin, D.; Yu, T.; Zaidi, A.; Mutuma, M.; Jha, R.; Awadallah, A.~H.;
  Çelikyilmaz, A.; Liu, Y.; Qiu, X.; and Radev, D. 2021.
\newblock QMSum: A New Benchmark for Query-based Multi-domain Meeting
  Summarization.
\newblock In \emph{NAACL}.

\bibitem[{Zhu et~al.(2021{\natexlab{a}})Zhu, Hinthorn, Xu, Zeng, Zeng, Huang,
  and Jiang}]{zhu-etal-2021-enhancing}
Zhu, C.; Hinthorn, W.; Xu, R.; Zeng, Q.; Zeng, M.; Huang, X.; and Jiang, M.
  2021{\natexlab{a}}.
\newblock Enhancing Factual Consistency of Abstractive Summarization.
\newblock In \emph{Proceedings of the 2021 Conference of the North American
  Chapter of the Association for Computational Linguistics: Human Language
  Technologies}, 718--733. Online: Association for Computational Linguistics.

\bibitem[{Zhu et~al.(2021{\natexlab{b}})Zhu, Liu, Mei, and
  Zeng}]{zhu-etal-2021-mediasum}
Zhu, C.; Liu, Y.; Mei, J.; and Zeng, M. 2021{\natexlab{b}}.
\newblock {M}edia{S}um: A Large-scale Media Interview Dataset for Dialogue
  Summarization.
\newblock In \emph{Proceedings of the 2021 Conference of the North American
  Chapter of the Association for Computational Linguistics: Human Language
  Technologies}, 5927--5934. Online: Association for Computational Linguistics.

\bibitem[{Zou et~al.(2021{\natexlab{a}})Zou, Lin, Zhao, Kang, Jiang, Sun,
  Zhang, Huang, and Liu}]{Zou2021UnsupervisedSF}
Zou, Y.; Lin, J.; Zhao, L.; Kang, Y.; Jiang, Z.; Sun, C.; Zhang, Q.; Huang, X.;
  and Liu, X. 2021{\natexlab{a}}.
\newblock Unsupervised Summarization for Chat Logs with Topic-Oriented Ranking
  and Context-Aware Auto-Encoders.
\newblock In \emph{AAAI}.

\bibitem[{Zou et~al.(2020)Zou, Zhao, Kang, Lin, Peng, Jiang, Sun, Zhang, Huang,
  and Liu}]{zou2020topic}
Zou, Y.; Zhao, L.; Kang, Y.; Lin, J.; Peng, M.; Jiang, Z.; Sun, C.; Zhang, Q.;
  Huang, X.; and Liu, X. 2020.
\newblock Topic-Oriented Spoken Dialogue Summarization for Customer Service
  with Saliency-Aware Topic Modeling.
\newblock \emph{arXiv preprint arXiv:2012.07311}.

\bibitem[{Zou et~al.(2021{\natexlab{b}})Zou, Zhao, Kang, Lin, Peng, Jiang, Sun,
  Zhang, Huang, and Liu}]{Zou2021TopicOrientedSD}
Zou, Y.; Zhao, L.; Kang, Y.; Lin, J.; Peng, M.; Jiang, Z.; Sun, C.; Zhang, Q.;
  Huang, X.; and Liu, X. 2021{\natexlab{b}}.
\newblock Topic-Oriented Spoken Dialogue Summarization for Customer Service
  with Saliency-Aware Topic Modeling.
\newblock In \emph{AAAI}.

\end{thebibliography}

\end{document}

\maketitle

\appendix
\section{Baseline Description}
In this section, we describe baselines in detail.

\textbf{Oracle}: This model is used to obtain an oracle with a greedy algorithm similar to \citet{10.5555/3298483.3298681}, which treats the utterances that maximize the ROUGE-2 as a summary.

\textbf{Lead-3}: This model is commonly used in the news summarization task, which treats the first three utterances of the document as a summary.

\textbf{BertExt}: This model is proposed by \citet{liu-lapata-2019-text}, which is an extractive model whose parameters are initialized with BERT.

\textbf{BertAbs}: This model is proposed by \citet{liu-lapata-2019-text}, which is an abstractive model with encoder initialized with BERT and trained with a different optimizer than its decoder.

\textbf{BART}: This model is proposed by \citet{Lewis2020BARTDS}, which is a state-of-the-art abstractive summarization model pre-trained with a denoising autoencoding objective.

\section{Implementation Details}
Our methods are implemented with PyTorch \cite{NEURIPS2019_bdbca288} and are based on both BertAbs \cite{liu-lapata-2019-text} and BART \cite{Lewis2020BARTDS}. For parameters in the original pre-trained models, we followed the default hyperparameter settings to train our summarizers. For our model built on BertAbs, there are 13 encoding layers, with the top layer randomly initialized and separately trained between the two encoders. For our model built on BART, there are 24 encoding layers, with the top layer initialized with pre-trained parameters yet separately trained between the two encoders. The first cross-attention block of the decoder is randomly initialized whereas the second cross-attention block is initialized with pre-trained parameters. Besides, The TRADE \cite{wu2019transferable} is used to predict dialogue state information during test time. Unless otherwise stated, we use oracle extractions at training time.

\section{An Example of User Goal in MultiWOZ}
\begin{figure*}[t]
\centering{
\includegraphics[scale=0.8]{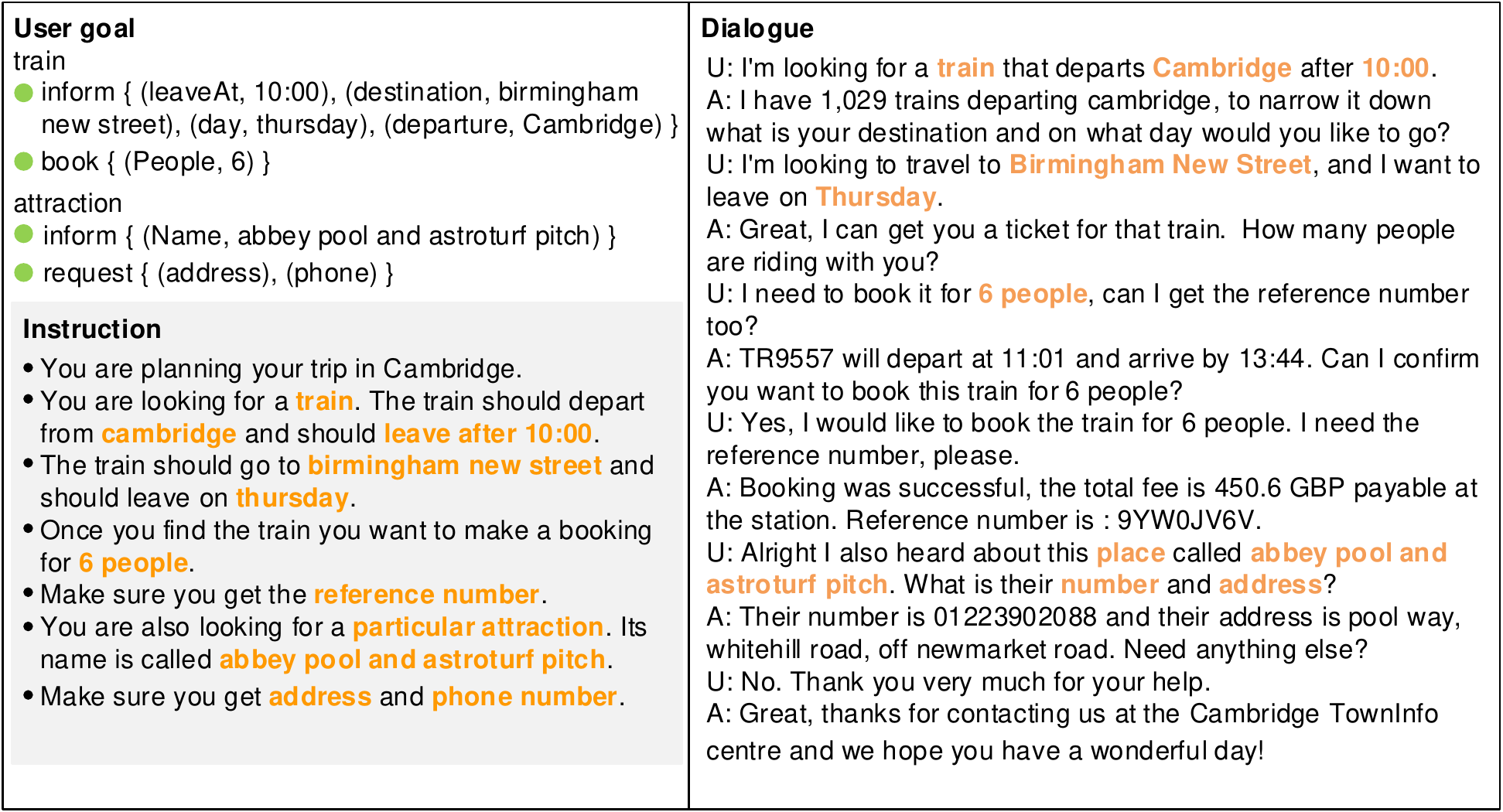}}
% \vspace{-0.3cm}
\caption{An example of a user goal and its explanation corresponding with the real dialogue.}
\label{goal_sample}
% \vspace{-0.3cm}
\end{figure*}

As we mention in the main paper, MultiWOZ uses a Wizard-of-Oz framework (WOZ) of human-to-human data collection. The Wizard-of-Oz framework (WOZ) is firstly proposed as an iterative approach to improve user experiences when designing a conversational system. The goal of WOZ data collection is to log down the conversation for future system development. To overcome the need of relying on the data collection to a small set of trusted workers, the collection set-up of MultiWOZ is designed to provide an easy-to-operate system interface for the Wizards and easy-to-follow goals for the users. This results in a bigger diversity and semantical richness of the collected data. The domain of a task-oriented dialogue system is often defined by an ontology, a structured representation of the back-end database. The ontology defines all entity attributes called slots and all possible values for each slot. In general, the slots may be divided into informable slots and requestable slots. Informable slots are attributes that allow the user to constrain the search (e.g., area or price range). Requestable slots represent additional information the users can request about a given entity (e.g., phone number). Based on a given ontology spanning several domains, a task template is created for each task through random sampling. This results in single and multi-domain dialogue scenarios and domain-specific constraints were generated. We display an example of a user goal and its explanation corresponding with the real dialogue in Figure \ref{goal_sample}.

\begin{figure*}[t]
\centering{
\includegraphics[scale=0.6]{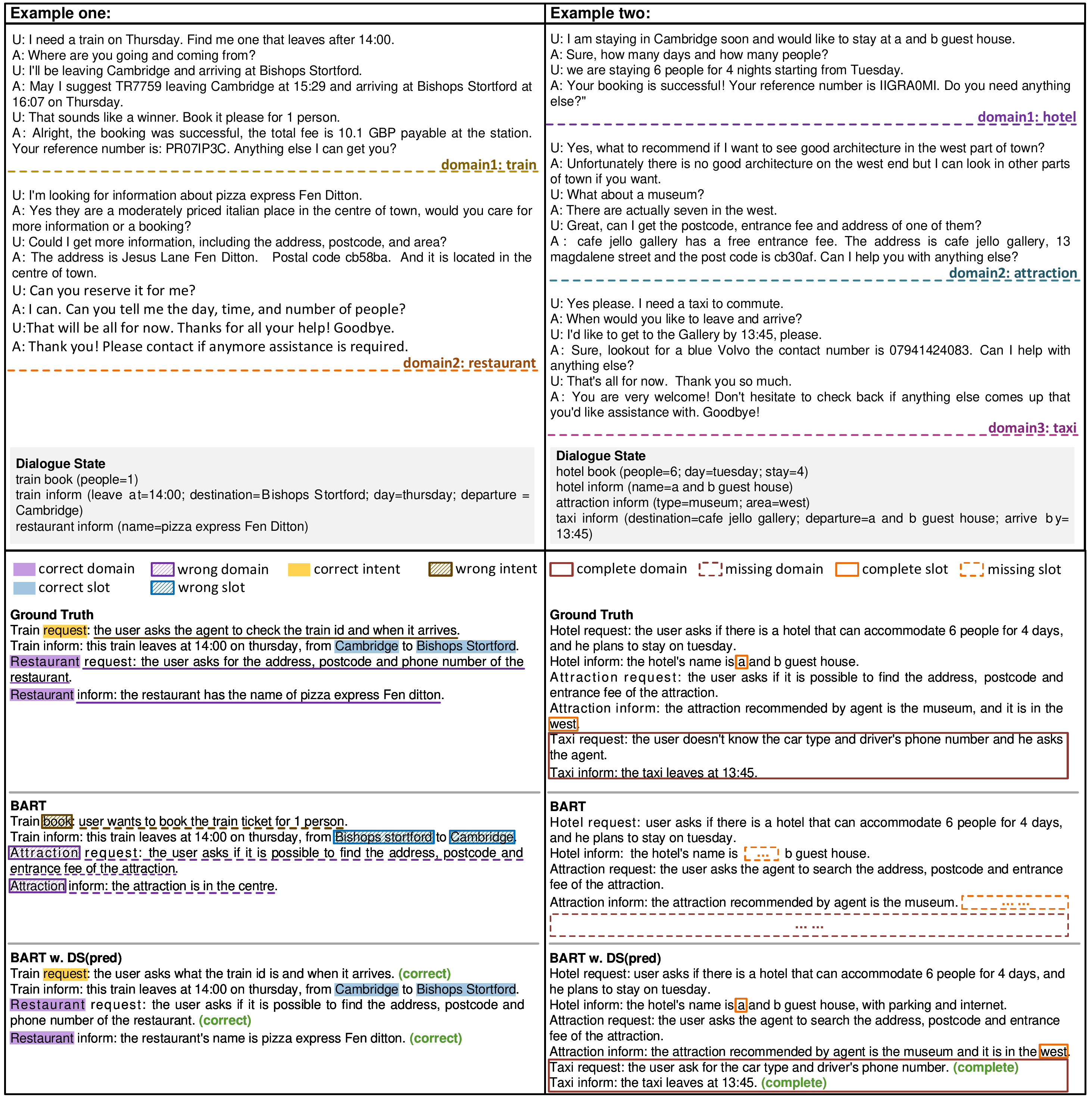}}
% \vspace{-0.3cm}
\caption{A case study for two examples from the TODSum dataset.}
\label{case_study}
% \vspace{-0.3cm}
\end{figure*}

\section{Case Study}
In this section, we show Figure \ref{case_study} for case study. The detailed introduction is in the main paper.

% \section{Acknowledgments}

\clearpage
\bibliography{aaai22.bib}